\documentclass[graybox,a4paper]{svmult}
\usepackage{geometry}
\usepackage[load=prefixed]{siunitx}
\usepackage{mathptmx}       
\usepackage{helvet}         
\usepackage{courier}        
\usepackage{makeidx}         
\usepackage{graphicx}        
\usepackage{multicol}        
\usepackage[bottom]{footmisc}
\usepackage[font=footnotesize, skip=5pt]{caption}
\usepackage{xcolor}
\usepackage{booktabs}
\usepackage{dcolumn}
\usepackage[load=prefixed]{siunitx}
\usepackage{layouts}  

\newcommand\tab[1][0.5cm]{\hspace*{#1}}
\newcolumntype{P}[1]{>{\centering\arraybackslash}p{#1}}

\newcommand{\secref}[1]{Sec.~\ref{#1}}

\newcommand{\eqref}[1]{Eq.~(\ref{#1})}
\newcommand{\figref}[1]{Fig.~\ref{#1}}
\newcommand{\tabref}[1]{Tab.~\ref{#1}}

\usepackage{titlesec}
\titlespacing*{\section}{0pt}{1.2\baselineskip}{\baselineskip}
\titlespacing*{\subsection}{0pt}{1.2\baselineskip}{\baselineskip}

\makeindex             

\begin{document}

\title*{Vision-Based Autonomous UAV Navigation and Landing for Urban Search and Rescue}
\author{Mayank Mittal\inst{1,2} \and Rohit Mohan\inst{2} \and Wolfram Burgard\inst{2,3} \and Abhinav Valada\inst{2}}
\institute{\inst{1}Department of Mechanical Engineering, ETH Zurich, Switzerland,\\
\inst{2}Department of Computer Science, University of Freiburg, Germany,\\
\inst{3}Toyota Research Institute, Los Altos, USA}
%
%
\maketitle

\abstract{Unmanned Aerial Vehicles (UAVs) equipped with bioradars are a life-saving technology that can enable identification of survivors under collapsed buildings in the aftermath of natural disasters such as earthquakes or gas explosions. However, these UAVs have to be able to autonomously navigate in disaster struck environments and land on debris piles in order to accurately locate the survivors. This problem is extremely challenging as pre-existing maps cannot be leveraged for navigation due to structural changes that may have occurred. Furthermore, existing landing site detection algorithms are not suitable to identify safe landing regions on debris piles. In this work, we present a computationally efficient system for autonomous UAV navigation and landing that does not require any prior knowledge about the environment. We propose a novel landing site detection algorithm that computes costmaps based on several hazard factors including terrain flatness, steepness, depth accuracy, and energy consumption information. We also introduce a first-of-a-kind synthetic dataset of over 1.2 million images of collapsed buildings with groundtruth depth, surface normals, semantics and camera pose information. We demonstrate the efficacy of our system using experiments from a city scale hyperrealistic simulation environment and in real-world scenarios with collapsed buildings.}

\section{Introduction}
\label{sec:Introduction}

Urban Search and Rescue (USAR) missions for finding victims in collapsed buildings is an extremely time-critical and dangerous task. There are several triggers for buildings to collapse including gas explosions, fires as well as natural disasters such as storms and earthquakes. The current paradigm followed during the response and recovery phase of the disaster management cycle is to first conduct a manual inspection of the damaged structures by the disaster response teams, followed by actions to search for victims using bioradars and thermal cameras. However, there are often inaccessible areas that may take anywhere from a few hours to days to reach. This not only endangers the lives of the trapped victims but also those of the rescue team due to the inherent instability of the rubble piles in such places.

\begin{figure}[t]
\centering
\setlength{\tabcolsep}{0.2cm}
\begin{tabular}{P{5.5cm} P{5.5cm}}
\includegraphics[width=\linewidth]{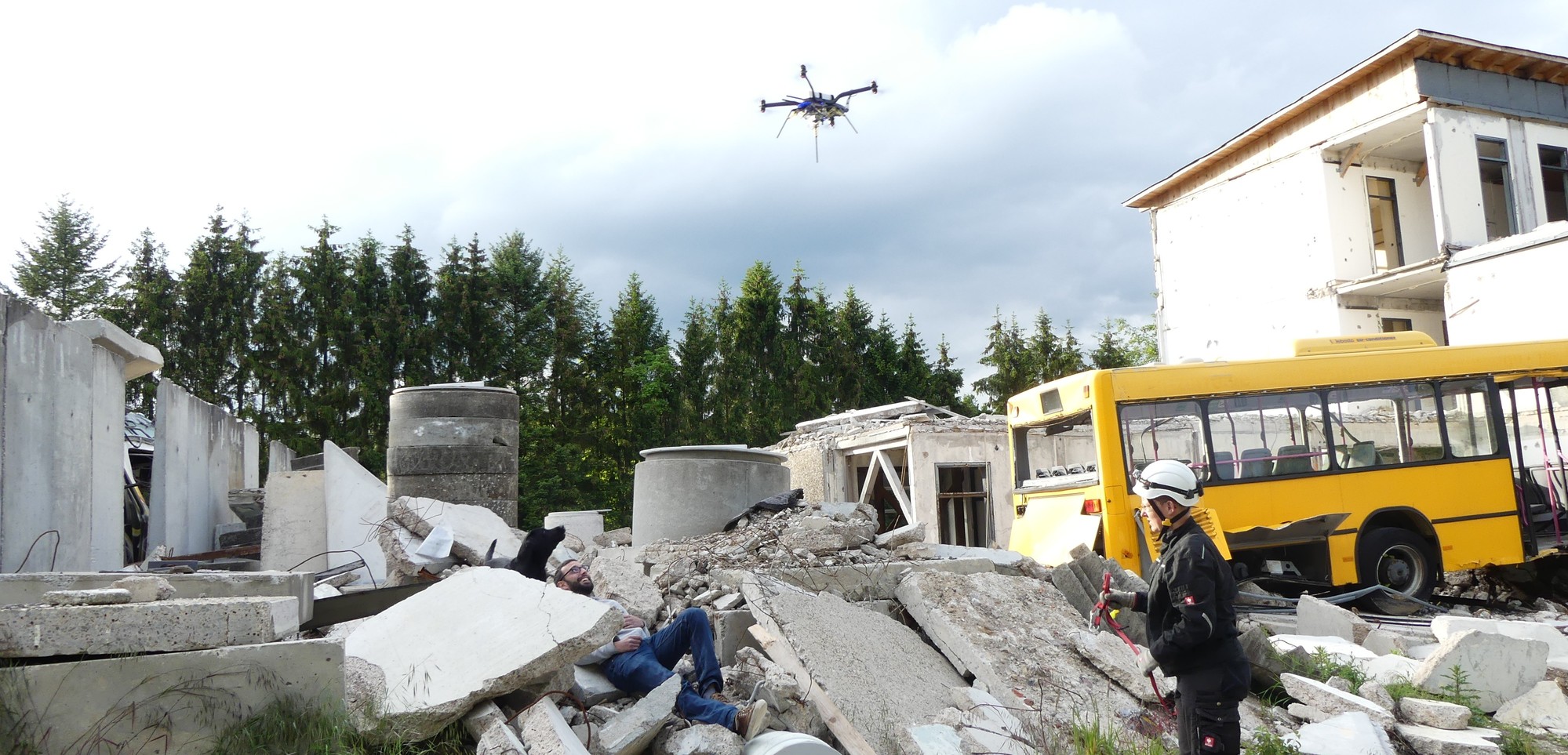} & 
\includegraphics[width=\linewidth]{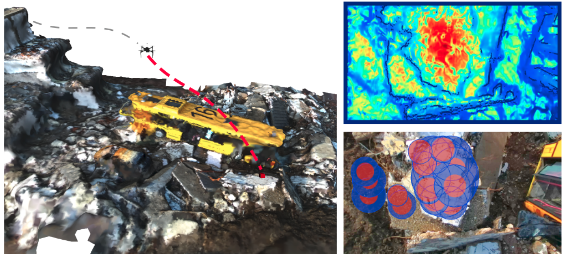} \\
\end{tabular}
\caption{\textit{Left}: Our UAV autonomously navigating in an environment with collapsed buildings attempting to identify trapped survivors. \textit{Middle}: A 3D volumetric reconstruction of the scene, additionally illustrating the minimum-jerk trajectory to the detected safe landing site. \textit{Right-Top}: The corresponding full costmap computed by our landing site detection algorithm. \textit{Right-Bottom}: Corresponding dense candidate landing sites that were detected projected onto the local 3D map.}
\label{fig:landing_cover}
\vspace{-0.5cm}
\end{figure}

These factors have increased the demand for employing UAVs for reconnaissance operations due to their agile maneuverability, fast deployment and ability to collect data at high temporal frequencies. Most UAVs that are employed in USAR operations today have limited capabilities and they need to be manually teleoperated. Training a human operator can take several months. Moreover, teleoperating UAVs in post-disaster scenarios is extremely challenging as almost no prior knowledge about the environment such as digital surface maps can be leveraged and a stable radio link cannot be guaranteed. In order to maximize the utility of UAVs for such safety-critical USAR missions, they should be able to autonomously operate with minimal human intervention.

Typically, in addition to optical sensors such as thermal cameras, ground penetrating radars such as bioradars can be used for bioradiolocation. These sensors can detect movements in the internal organs of humans such as the lungs and heart through thick layers of structures. Recently, our collaborators have miniaturized a bioradar~\cite{shi2017design,valada18irosws} which can be mounted on a small-sized UAV. However, due to a large difference in the electromagnetic impedance between the air and collapsed structures, the antenna of the bioradar should be in contact with a surface in order to obtain accurate measurements. Therefore, the UAV should be capable of reliably detecting safe landing sites in collapsed structures and autonomously perform the landing maneuver. Most existing work on landing site detection utilize specific markers or patterns that can be identified by the UAV, such as the alphabet 'H' on helipads or fiducial markers. These approaches require the landing site to be predetermined and are not employable in unstructured or unknown environments. Other existing approaches that only use planarity constraints~\cite{bosch2006autonomous} are insufficient for our application as the UAV is required to land on rubble piles. Due to the nature of these collapsed structures, the algorithm should be able to handle a multitude of visual terrain features and natural or man-made obstacles that the UAV might encounter in the post-disaster struck environment. Moreover, a critical requirement being that the landing site detection algorithm has to run online on a low-power embedded computer along with other autonomy packages for state estimation, exploration, and mapping.\looseness=-1

In this paper, we present an efficient vision-based autonomous navigation and landing system for UAVs equipped with bioradars that conduct USAR missions in post-disaster struck environments. Our autonomy system enables the UAV to reliably localize itself, build three-dimensional occupancy maps, plan collision-free trajectories and autonomously explore highly complex unstructured environments while reconstructing a three-dimensional textured mesh of the scene. As the main goal of our UAV is to land on rubble piles and locate trapped victims using the bioradar, we propose a novel landing site detection algorithm that identifies safe landing regions on collapsed buildings without any prior assumptions about the environment (\figref{fig:landing_cover}). Our proposed algorithm assesses the risk of a landing site by evaluating the flatness and inclination of the terrain, the confidence of the depth information and the energy required to land on the detected site. We first estimate a set of dense candidate landing sites locally by computing a weighted sum of the costmaps for each of the aforementioned hazard factors. This is followed by optimizing over the global area explored by the UAV using a clustering-based approach. Once the final landing site has been determined, the system computes a minimum-jerk trajectory considering the nearby obstacles as well as the UAV dynamics and executes the landing maneuver. Our proposed algorithm can also be employed in other scenarios, such as for landing in a planned or an emergency situation. More notably, our entire navigation and landing system runs online on the UAV on a low-power embedded computer.

To the best of our knowledge, there is no publicly available dataset on collapsed buildings with aerial RGB images, depth and camera pose information. In order to facilitate comparison to existing techniques and to encourage future work in this direction, we introduce a new large-scale synthetic dataset gathered in our city-scale hyperrealistic simulated environment which resembles a region struck by a natural disaster. Our AutoLand dataset consists of over 1.2 Million RGB images, groundtruth depth, surface normals, pixel-level semantic segmentation labels, and camera pose information. We evaluate our proposed system using extensive experiments on our synthetic dataset as well as from five days of operation in real-world environments with collapsed buildings. The work that we present in this paper is the first autonomous UAV system capable of landing on collapsed structures and identifying trapped victims in the aftermath of natural disasters.

\section{Related Work}
\label{sec:relatedWork}

In the last decade, there has been an increasing interest to employ UAVs for several civilian and scientific applications, such as surveying and mapping, building inspection and rescue operations. In this section, we focus on the literature closely related to our work on autonomous UAVs for USAR missions and landing sites detection.

One of the pressing challenges for using robots in USAR operations is to make the execution efficient in order to find more survivors and provide timely information to first responders. A variety of ground and aerial robots have been proposed in the past for this task~\cite{Murphy2008}. Goodrich \textit{et~al.}~\cite{goodrich08jfr} motivate how vision-based UAVs can assist in wildlife search and rescue operations for tasks such as area coverage. Tomic \textit{et~al.}~\cite{tomic12ram} present a fully autonomous UAV for USAR missions which can perform real-time navigation. A large variety of work exists on navigation in complex GPS denied~\cite{Nieuwenhuisen2016} and unknown environments~\cite{bachrach90}. Such a system typically runs a simultaneous localization and mapping (SLAM) algorithm for navigation and exploration. Leveraging different sensor modalities such as a visual-inertial setup~\cite{bloesch2015robust} helps in improving the performance in these tasks.

Conventional approaches for landing site detection from aerial images have employed fiducial markers and several types of such markers have been proposed for this purpose including point markers, H-shaped markers, and square markers. The markers are detected on the basis of their color or geometry using classical image features and then the relative pose of the UAV is estimated from these extracted feature points. While the usage of markers for landing purposes is reliable and efficient, they are usable only when the desired landing spots are known in advance such as for landing on ship decks~\cite{grocholskyrobust} or a moving mobile robot platform~\cite{falanga2017vision}.

More relevant to our work are the approaches that estimate safe landing sites in unknown or unstructured environments. A Lidar-based approach~\cite{johnson2002lidar} is prohibitively expensive for USAR missions due to the large payload and power requirements. Therefore, a variety of vision-based approaches have been proposed in the last decade. Desaraju~\textit{et~al.}~\cite{desaraju2014vision} employ an active perception strategy utilizing Gaussian processes to estimate feasible rooftop landing sites along with the landing site uncertainty as assessed by the vision system. While being very powerful, Gaussian Processes, unfortunately, are computationally expensive. On the other hand, Forster~\textit{et~al.}~\cite{forster2015continuous} propose a monocular camera-based system that builds a 2.5D probabilistic elevation map from which landing sites are detected over regions that are flat. Similarly, a stereo vision-based landing site search algorithm is presented by Park \textit{et~al.}~\cite{park2015landing} which computes a performance index for landing by considering the depth, flatness, and energy required to reach a specific site. Most recently, Hinzmann~\textit{et~al.}~\cite{hinzmann2018free} presented a landing site detection algorithm for autonomous planes that employs a binary random forest classifier to select regions with grass, followed by a 3D reconstruction of the most promising regions from which hazards factors such as terrain roughness, slope and the proximity to obstacles are computed to determine the landing region. Since catastrophe-struck areas do not share the same terrain properties, their approach is less suited for our application. 

In contrast to the aforementioned techniques, the approach presented in this paper detects safe landing sites on rubble piles of collapsed buildings. It employs fine-grained terrain assessment considering a wide range of hazard factors at the pixel-level. By first estimating dense candidate landing sites in a local region, followed by a global refinement, our approach can run online along with other state estimation and mapping processes on a single low-power embedded computer on our UAV.

\section{Technical Approach}
\label{sec:technicalApproach}

\begin{figure*}
\centering
\includegraphics[width=\linewidth]{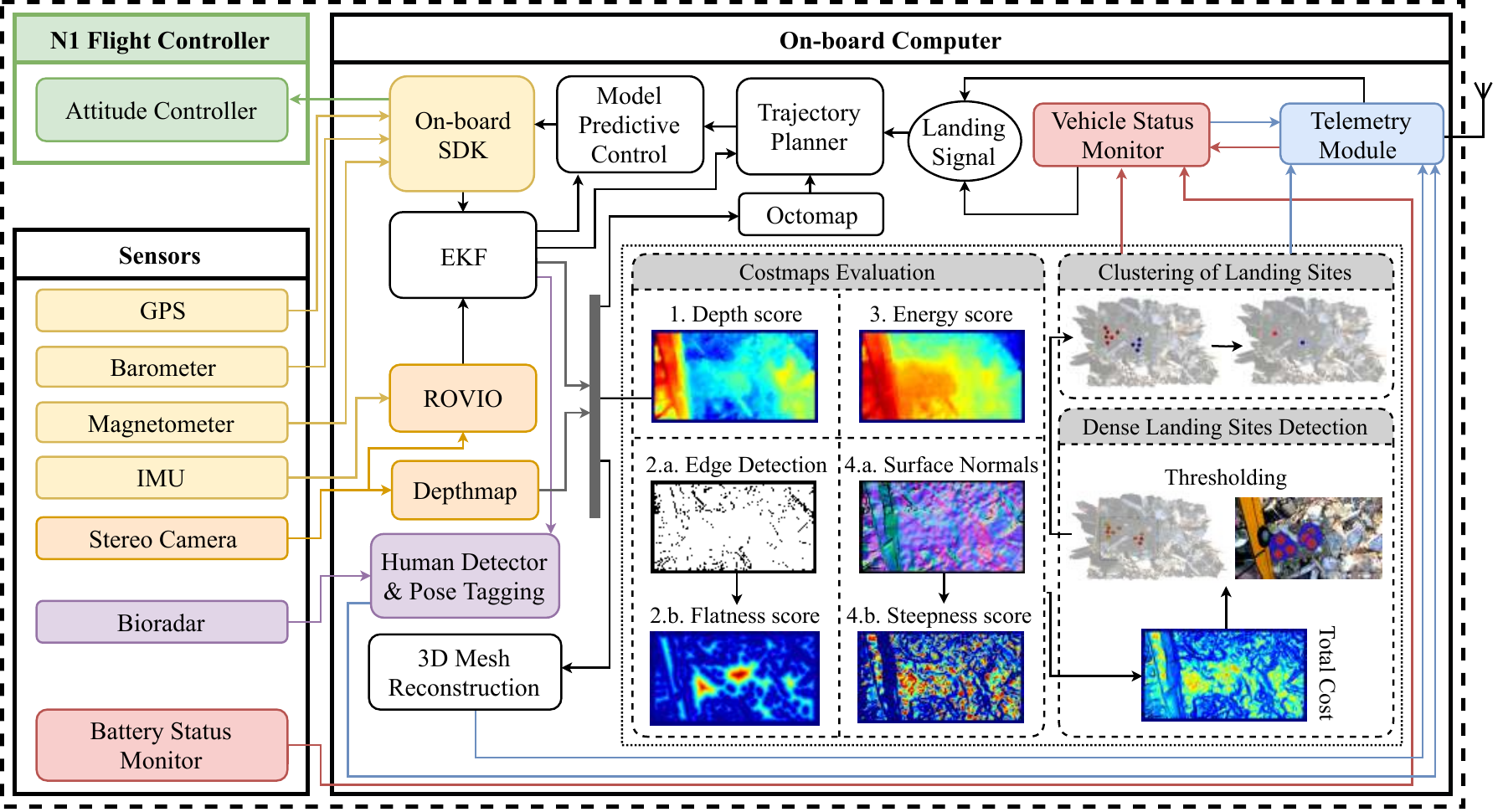}
\caption{Overview of our autonomous navigation and landing system. We use the DJI M100 quadrotor with the N1 flight controller, the NVIDIA Jetson TX2 for computation, and a ZED stereo camera for acquiring depth information. All our mapping, localization and landing site detection algorithms run online on the Jetson TX2.}
\label{fig:overview_system}
\vspace{-0.5cm}
\end{figure*}

In this section, we first briefly describe the architecture of our autonomous navigation and landing system shown in \figref{fig:overview_system}, followed by detailed descriptions of each of the constituting components. We estimate the pose of the UAV by fusing raw data from dead-reckoning sensors, namely GPS and the pose obtained from a visual-inertial odometry framework as described in~\secref{sec:stateMapExplore}. In order to safely navigate in a disaster struck environment, the UAV should first explore the area while building an occupancy map using depth images acquired by the stereo camera. We use a receding horizon next-best-view planner for efficient exploration. Simultaneously, we build two different 3D representations of the environment - an Octomap~\cite{hornung13auro} which serves as a light voxel-based internal map for trajectory planning and a 3D mesh reconstruction~\cite{oleynikova2017voxblox} which is transmitted to the ground station for analysis and rescue planning. Additionally, the UAV transmits the vehicle status information as well as landing sites that are detected in the explored area.

In order to detect safe landing sites, we employ our proposed protocol detailed in~\secref{sec:landingDetection} which consists of three stages. In the first stage, we evaluate the costmaps for various metrics including terrain flatness and steepness, depth accuracy, and energy consumption information in the local camera frame. We then infer a set of dense candidate landing sites from the combined costmaps, followed by employing a nearest neighbor filtering and clustering algorithm to obtain a sparser set of unique landing sites in the world frame. For the sake of generality, we express the situations for planned (normal operations) and forced landings (emergencies) using a common landing signal. In case of a planned operation, the operator selects a specific site to land from the list of candidate sites detected in the region. While, in case of emergencies, the UAV chooses a landing site based on whether it needs to land quickly (low battery) or at a location closest to the remote station (loss of communication signal). Once the landing signal is received, a minimum-jerk trajectory is planned to safely land at the selected site as described in \secref{sec:trajectoryEstimation}. Finally, waypoints indicating the planned trajectory are fed to a position-based model predictive controller which sends the actuation commands to the flight controller.

\subsection{State Estimation, Exploration, and Mapping}
\label{sec:stateMapExplore}

Accurate estimation of the pose of the UAV is an essential prerequisite for various modules including volumetric mapping, localization of detected humans from the bioradar measurements and for detecting landing sites using our proposed algorithm. Although GPS provides a straightforward solution for outdoor environments, it is highly unreliable in post-disaster scenarios. Therefore, we use the Robust Visual Inertial Odometry (ROVIO) \cite{bloesch2015robust}, a robot-centric visual-inertial state estimation framework that uses pixel intensity errors for detection of multi-level patch features, and tracks the detected features using an iterated Extended Kalman Filter (EKF). The ROVIO module takes the image stream from the downward-facing camera and measurements from the inertial measurement unit (IMU) mounted on the UAV as inputs and estimates the pose of the vehicle. Furthermore, we fuse this pose estimate with raw sensor data from other onboard sensors such as the GPS and the barometer using an EKF to avoid drift.

Exploring and building a map of an unknown environment consisting of collapsed buildings is one of the foremost tasks for the UAV during the search and rescue operation. The map is not only used for its navigation but also by the rescue team to remotely assess the situation. For exploration, we use a sampling-based planning algorithm which employs a receding horizon next-best-view scheme~\cite{bircher2016nbv}. The planner incrementally builds a random tree using the RRT algorithm and at each step, it chooses the branch that maximizes the unmapped volume that can be explored at the nodes along the branch. In order to create 3D volumetric maps, the UAV primary relies on the depth information from the stereo camera to sense the environment. We generate two different 3D map representations, an occupancy grid for planning and a textured 3D mesh for visualization. The two maps are generated at different resolutions since a high-resolution map is required for planning a human-lead rescue operation, while a low-resolution map enables faster trajectory planning of collision-free paths for the UAV navigation. We use Octomaps~\cite{hornung13auro} for the internal representation of the environment at a low-resolution (typically $0.5\meter$) and Voxblox~\cite{oleynikova2017voxblox} which is based on Truncated Signed Distance Fields (TSDFs) for generating a 3D textured mesh. This framework allows for dynamically growing maps by storing voxels in the form a hash table which makes accessing them more efficient compared to an octree. The reconstructed mesh is computed on the onboard NVIDIA Jetson TX2 and transmitted to the ground station for analysis by the rescue team.

\subsection{Landing Site Detection}
\label{sec:landingDetection}

\begin{figure*}
\centering
\includegraphics[width=\linewidth]{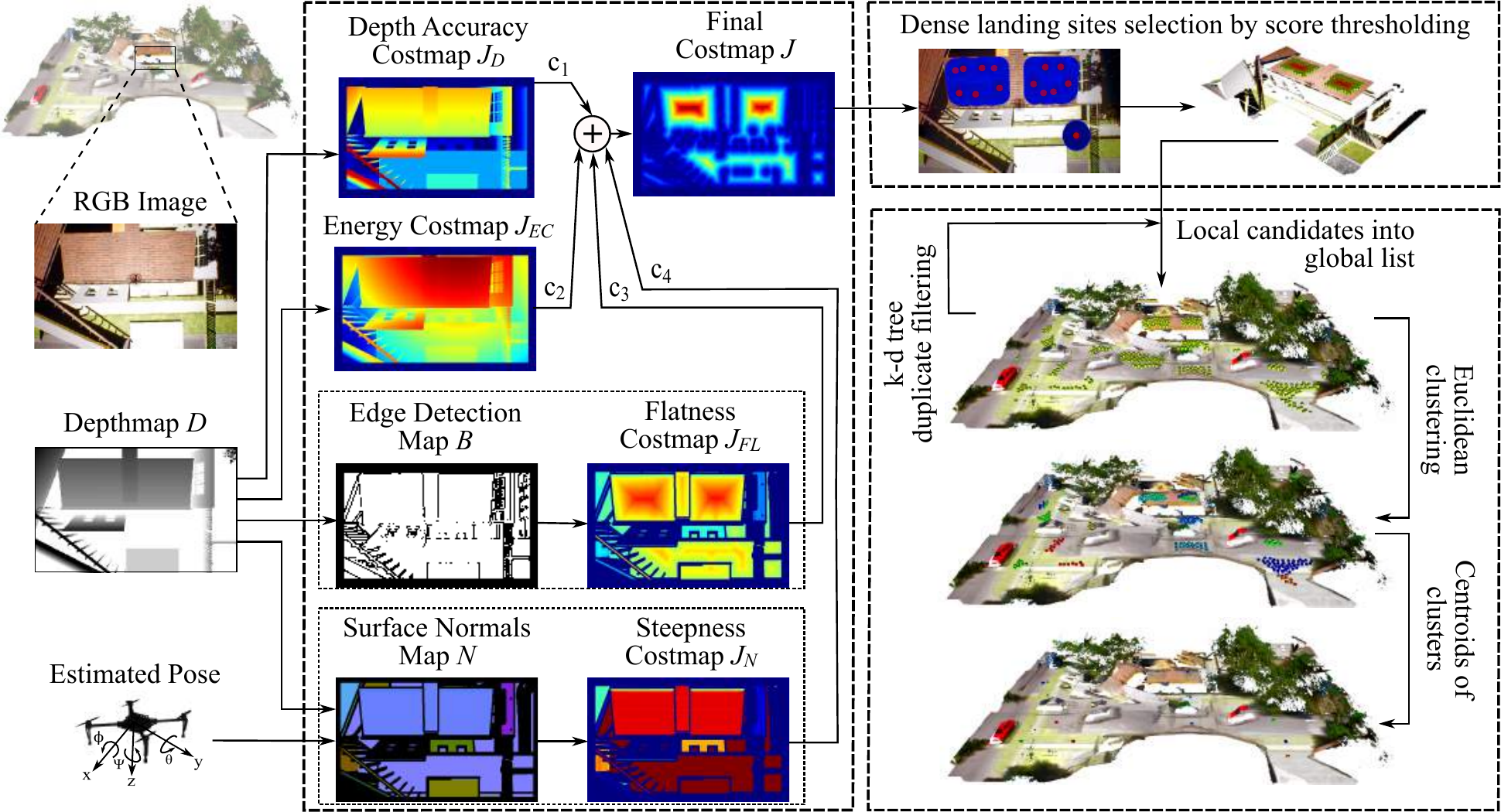}
\caption{Overview of our landing site detection algorithm. The figure illustrates the various costmaps for hazard estimation. \textit{Scale:} Red indicates highest assigned score while blue indicates the lowest score. The detected landing sites are projected on to a volumetric 3D reconstructed mesh of the environment.}
\label{fig:landing_algo}
\vspace{-0.5cm}
\end{figure*}

The criteria for the selection of candidate landing sites is to locate regions from aerial imagery that are reasonably flat, within the range of the accepted slope, free of obstacles and large enough for the UAV to land on. Quantifying each of these requirements through a costmap makes our approach generic to the various structures that the UAV may encounter in the aftermath of catastrophic events. Using the estimated pose of the UAV and the depth map $D$ obtained from synchronized stereo images, the key costmaps computed in our algorithm are described as follows: 

{\parskip=5pt
\noindent\textbf{Confidence in Depth Information $J_{DE}$:} In~\cite{nguyen2012kinect}, Nguyen~\textit{et al.} empirically derive a noise model for an RGB-D sensor. According to this model, the variance of the axial noise distribution of a single ray measurement varies as a squared function of the depth measurement in the camera frame. Thus when the UAV is at high altitude, inaccurate depth information may be obtained for objects closer to the ground. In order to encode this confidence on the depth information, we evaluate a costmap $J_{DE}$ where the score assigned to each pixel $p=(x, y)$ is given by:}
\begin{equation}
J_{DE}(p) = - D(p)^2
\end{equation}
{\parskip=5pt
\noindent\textbf{Flatness Information $J_{FL}$:} From the depth map, the flatness of an area can be represented by the portion in the map having the same depth. The size of this area can be determined by the diameter of a circle inscribed in every portion of the scene present at the same elevation. By applying a Canny edge detector over the depthmap, we obtain a binary image where non-zero elements represent the edges for depth discontinuities. In order to find the inscribed circle, we apply a distance transform method which assigns a number to each pixel depending on the distance of that pixel to the nearest nonzero pixel in the binary edge map. For each pixel, $p=(x,y)$, the Euclidean distance transform of a binary image $B$, is defined as}
\begin{equation}
di(B, p) = min\Big\{\sqrt{(p-q)^T (p-q)} \Big{|} B(q) = 1 \Big\}
\end{equation}
Using this operator, we calculate the flatness score as
\begin{equation}
J_{FL}(p) = di(Canny(D), p).
\end{equation}
{\parskip=5pt
\noindent\textbf{Steepness Information $J_N$:} Another criterion to measure the quality of a landing site is based on the steepness of the area around the region. We quantify the steepness by computing the surface normals from the generated depth map. We generate a point cloud from the depth map in the camera frame. To account for the orientation of the UAV, we transform the point cloud to the local body-fixed frame of the UAV. Using the transformed depth information, we estimate the surface normals map $N$ using average 3D gradients algorithm. Compared to the covariance-based method, the average 3D gradients approach computes the normals significantly faster and the results are comparable to a 30-nearest neighborhood search in the covariance method~\cite{nakagawa3dcv}. For each pixel $p$, we evaluate the slope with respect to the $\mathbf{z}$-axis of the world frame (parallel to the gravity vector) as:}
\begin{equation}
	\theta = cos^{-1}(\hat{n}.\hat{z})
\end{equation}
where $\hat{n}$ is the computed surface normal direction. The steepness score for each pixel $p$ is then given by:
\begin{equation}
	J_N(p) = \exp\Bigg\{-\frac{\theta^2}{2\theta_{th}^2}\Bigg\}.
\end{equation}
We set $\theta_{th}$ to $15^o$ in this work, which is the maximum tolerable slope that our UAV can perch on safely.

{\parskip=5pt
\noindent\textbf{Energy Consumption Information $J_{EC}$:} Often, there are several flat areas where the UAV can potentially land. In some cases, it might be desirable to land on a site that will consume the least amount of energy to navigate to. In order to account for this factor, we compute the energy consumption required to follow a safe trajectory to a landing site at pixel $p$ as
\begin{equation}
J_{EC}(p) = \int_{t_{o}}^{t_{f}} P(t) dt
\end{equation}
where $t_0$ and $t_f$ are the time of flight for the path to reach the location at pixel $p$, and $P(t)$ is the instantaneous battery power. However, computing a costmap by evaluating this integral is a computationally expensive task since a trajectory for the UAV would need to be computed for each pixel. As shown in\cite{park2015landing}, the battery consumed by the UAV is directly related to the Euclidean distance between the UAV and the location of the point in 3D space. Following this approximation, we assign the computed values to each pixel to obtain the costmap $J_{EC}$.}

{\parskip=5pt
\noindent\textbf{Dense Landing Sites Detection:} 
After evaluating the individual costmaps, we perform min-max normalization over the depth accuracy, flatness and the energy consumption costmaps to scale their values to the same range and remove any biases due to unscaled measurements in the evaluated costs. We then take a weighted sum of the scores assigned to each pixel in their respective costmaps and calculate a final decision map $J$ given by
\begin{equation}
J = c_1 J_{DE} + c_2 J_{FL} + c_3 J_{N} + c_4 J_{EC}
\end{equation}
where $c_1$, $c_2$, $c_3$ and $c_4$ are weighting parameters for each map with the constraints $c_i \in [0,1]$, $\forall i \in \{1, 2, 3, 4\}$ and $c_1 + c_2 + c_3+ c_4 =1$.
The sites with scores above a certain threshold are considered as candidate landing sites. We perform further filtering of the landing site by evaluating whether the area available around each candidate site is large enough for a UAV to land on. We achieve this by comparing the flatness score of the site to the pixel-wise size of the UAV obtained by projecting the true size of the UAV on to the image plane using a pinhole camera model. Once the filtering  is performed, the filtered candidate sites are forwarded to the next stage for aggregation into a global list of detected sites.}

{\parskip=5pt
\noindent\textbf{Clustering of Landing Sites:} As we perform landing site detection on a frame-to-frame basis, a landing site detected in the current image frame might be detected again in future frames. In order to account for this repeatability, we maintain a global list of landing sites. For each candidate site, we use the depth information and the pose of the UAV to infer its 3D position in the world frame. We add the location of the new landing site to the global list only if it currently has no existing neighbors in the list within a small distance threshold. The global list uses a k-D tree. This data structure choice makes it efficient to find a neighboring landing site to an input candidate site.}

Since in large flat regions, several landing sites might be detected within proximity to each other. If the entire exhaustive list of detected sites is provided to the human operator, it reduces the reaction time in making a good decision during time-critical rescue situations. In order to overcome this problem, we apply an agglomerative hierarchical clustering algorithm over the global list of detected landing sites. Clusters are formed based on the Euclidean distance between two landing sites and the difference in their locations along the z-axis of the world frame (parallel to the gravity vector). This yields a sparse set of landing sites with each site location corresponding to the centroid of a cluster. The score assigned to each centroid is an average of the score assigned to each site present in that cluster. We set the distance threshold between two clusters as a factor of the size of our UAV.



\subsection{Landing Trajectory Estimation}
\label{sec:trajectoryEstimation}

When the human operator selects a landing site or when the on-board vehicle status monitor detects the need to land, the UAV plans a trajectory to the selected site and initiates the landing maneuver. To do so, we utilize a minimum-jerk trajectory generator~\cite{richter2016polynomial} with non-linear optimization~\cite{burri2015real}. The algorithm firsts find a collision-free path to the landing site using RRT*, followed by generating waypoints from the optimal path according to a line-of-sight technique. Using unconstrained nonlinear optimization, it generates minimum-snap polynomial trajectories while considering the minimal set of waypoints and a differential flat model of the quadrotor. This allows the UAV to travel in high-speed arcs in obstacle-free regions and ensures low velocities in tight places for minimum jerk around corners. 

\section{Experimental Results}
\label{sec:experiments}


We evaluate our system extensively in both simulated and real-world disaster scenarios with collapsed buildings. We created a city-scale simulation environment using the Unreal Engine consisting of collapsed buildings, damaged roads, debris and rubble piles, overturned cars, uprooted trees and several other features resembling a catastrophe struck environment. While, for real-world evaluations, we exhaustively performed experiments for five days of operation at the TCRH Training Center for Rescue in Germany, which spans an area of 60,000 square meters and consists of buildings with earthquake and fire damage.

\subsection{Experiments in Hyperrealistic Simulation}

\begin{figure}[t]
\centering
\footnotesize
\setlength{\tabcolsep}{0.15cm}
\begin{tabular}{P{2.5cm} P{2.5cm} P{2.5cm} P{2.5cm}}
RGB & Depth & Surface Normals & Semantics \\ 
    \includegraphics[width=\linewidth]{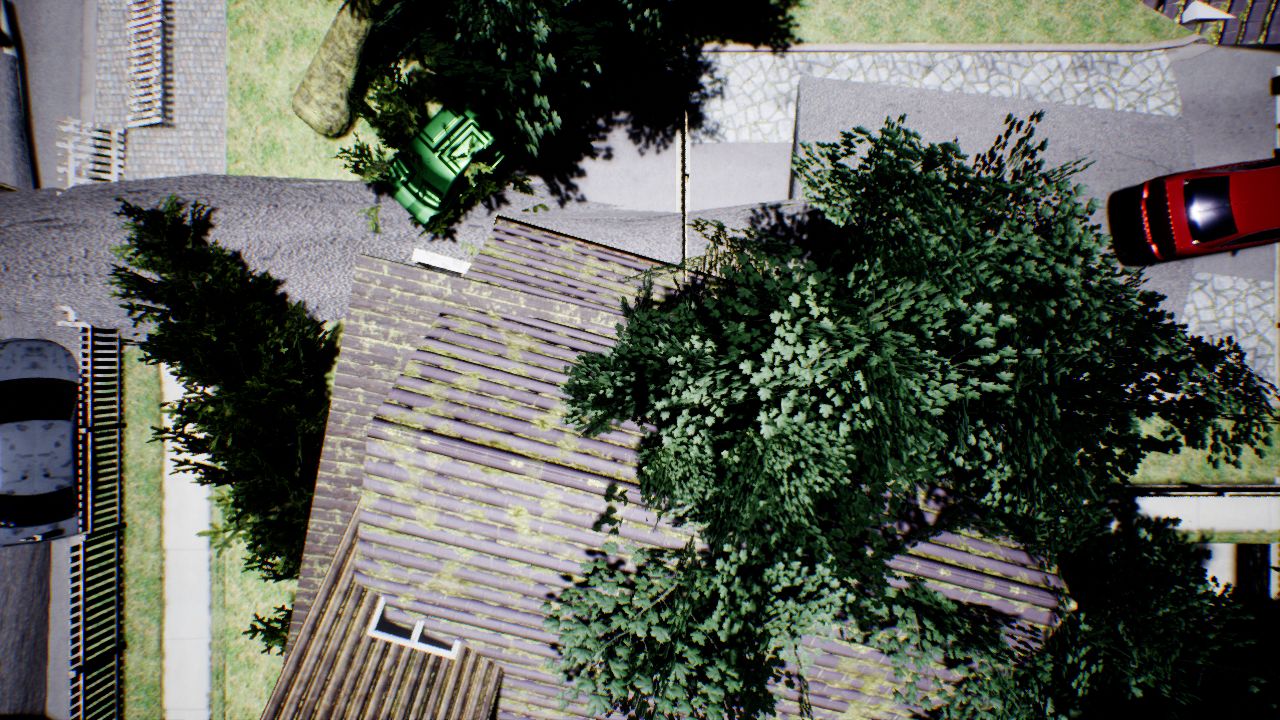} & 
    \includegraphics[width=\linewidth]{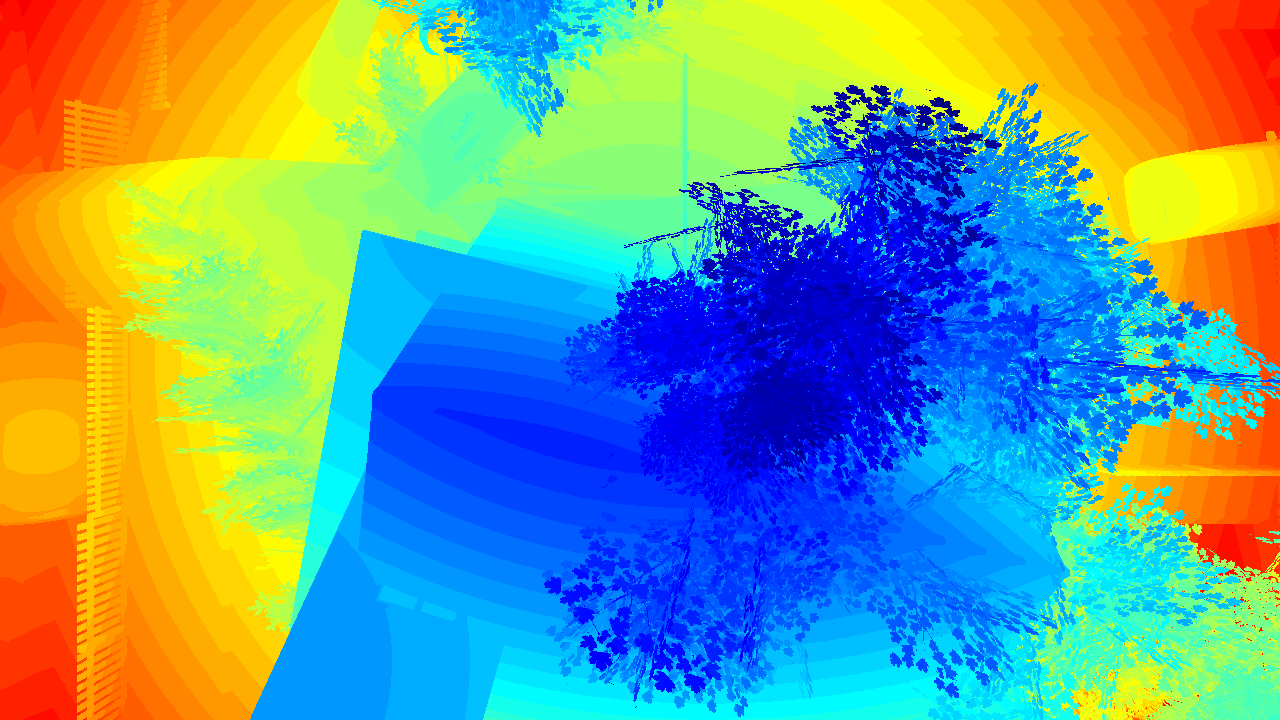} & 
    \includegraphics[width=\linewidth]{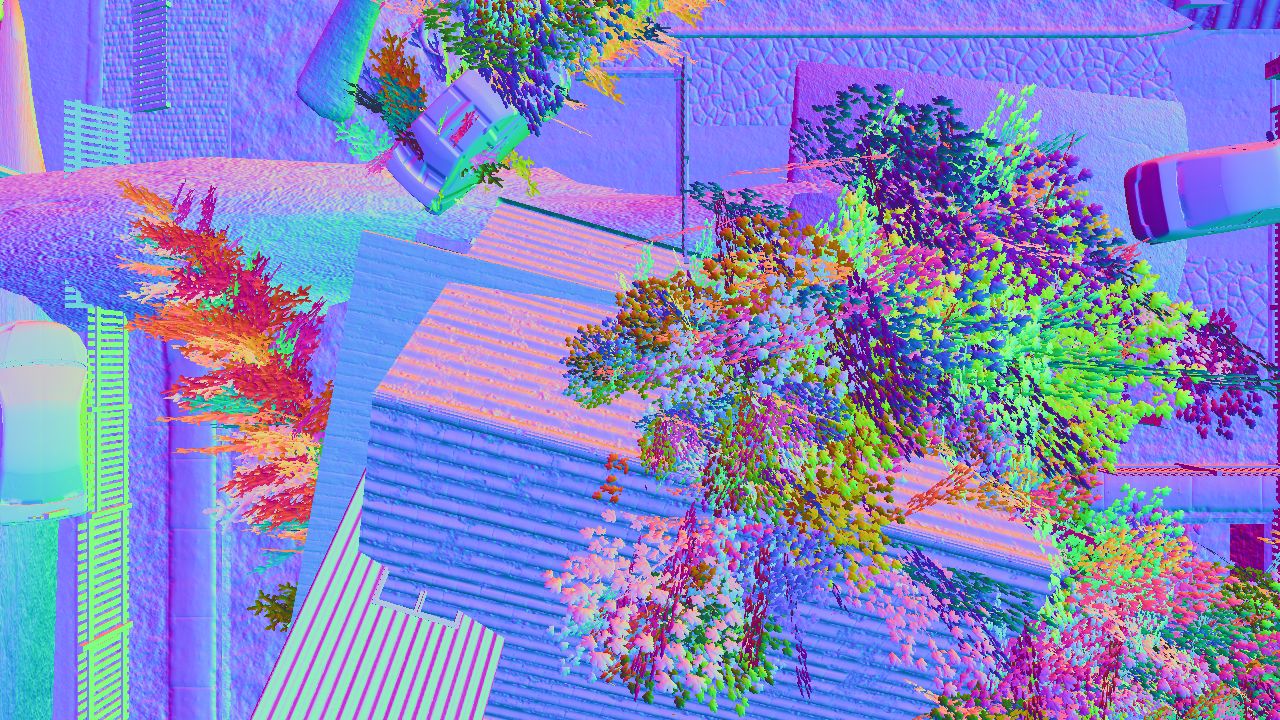} & 
    \includegraphics[width=\linewidth]{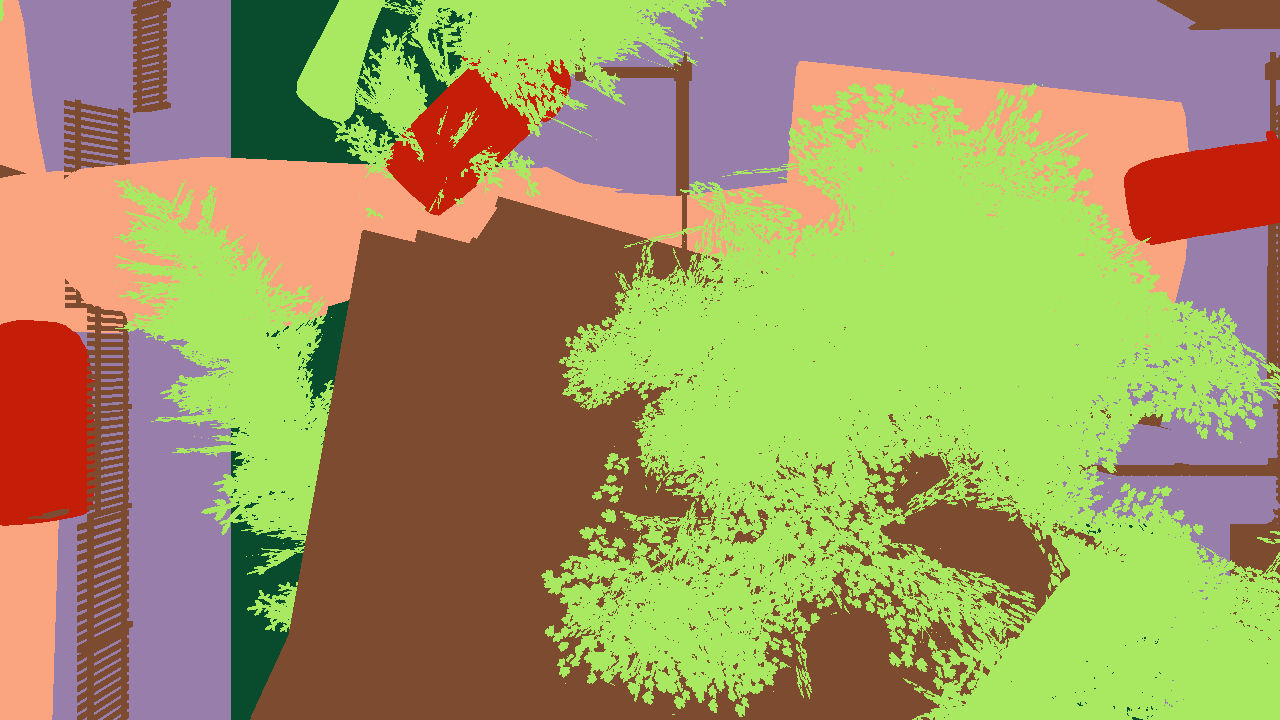} \\
    \includegraphics[width=\linewidth]{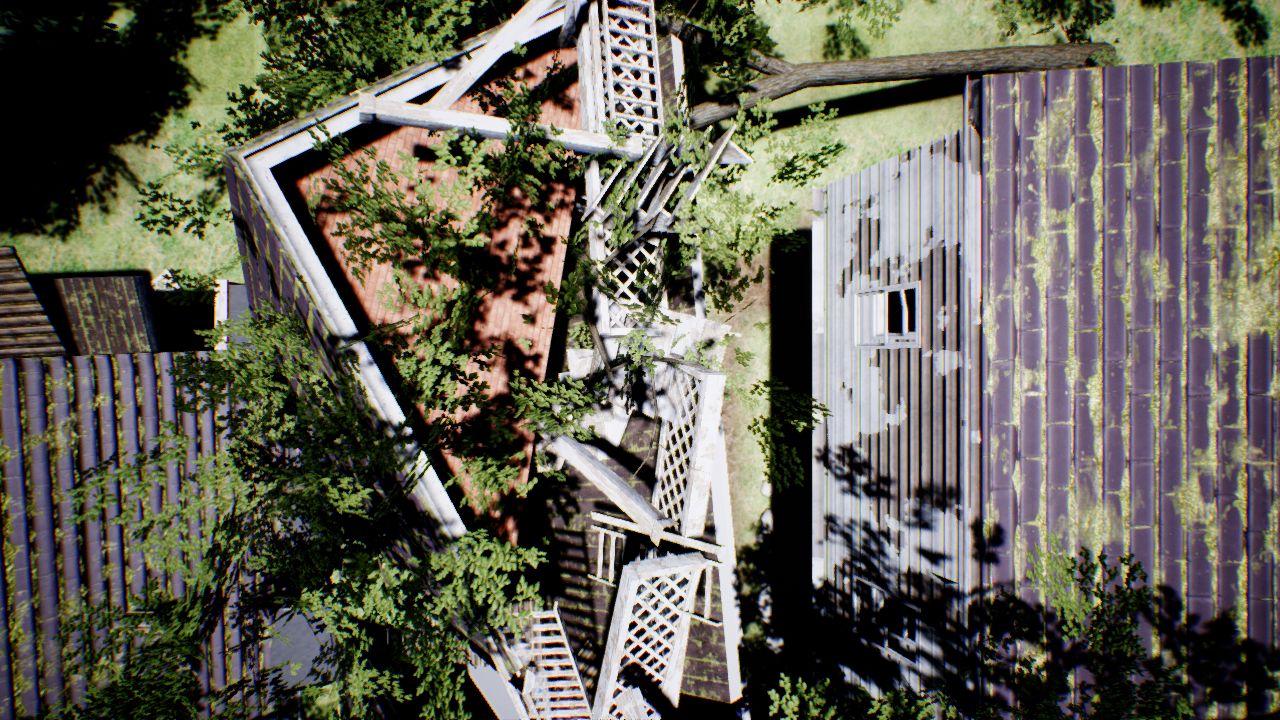} & 
    \includegraphics[width=\linewidth]{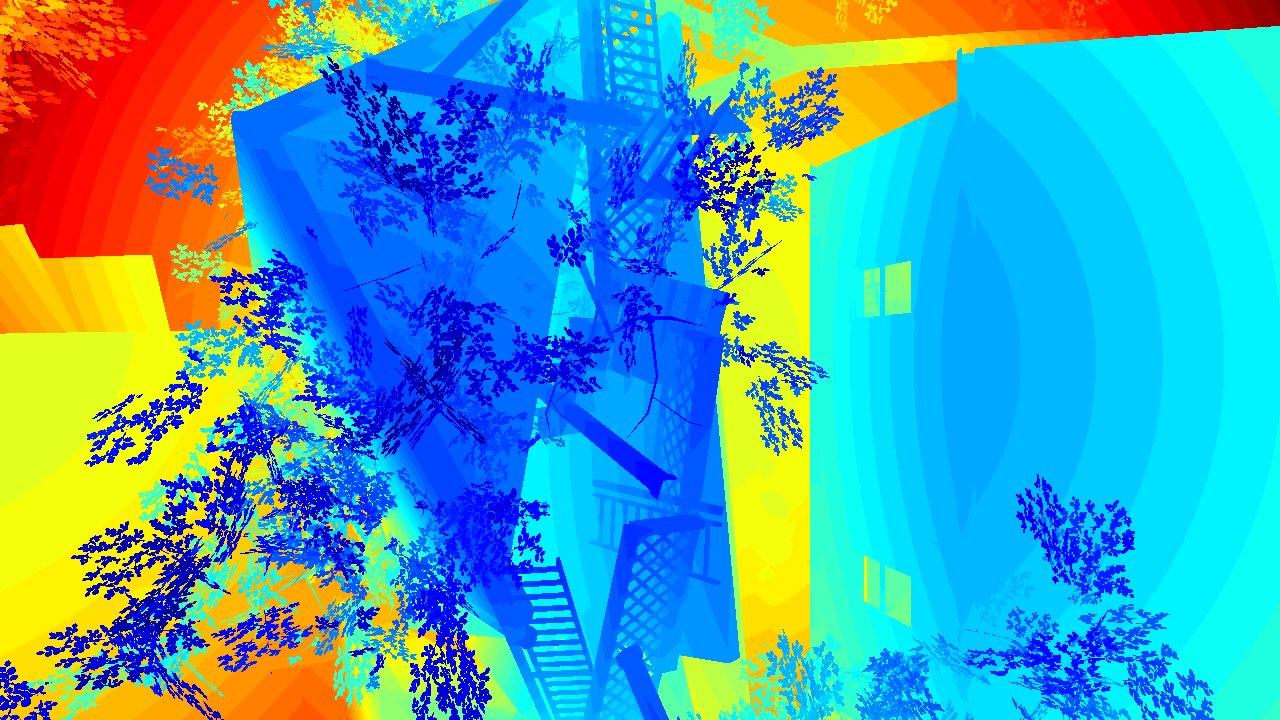} & 
    \includegraphics[width=\linewidth]{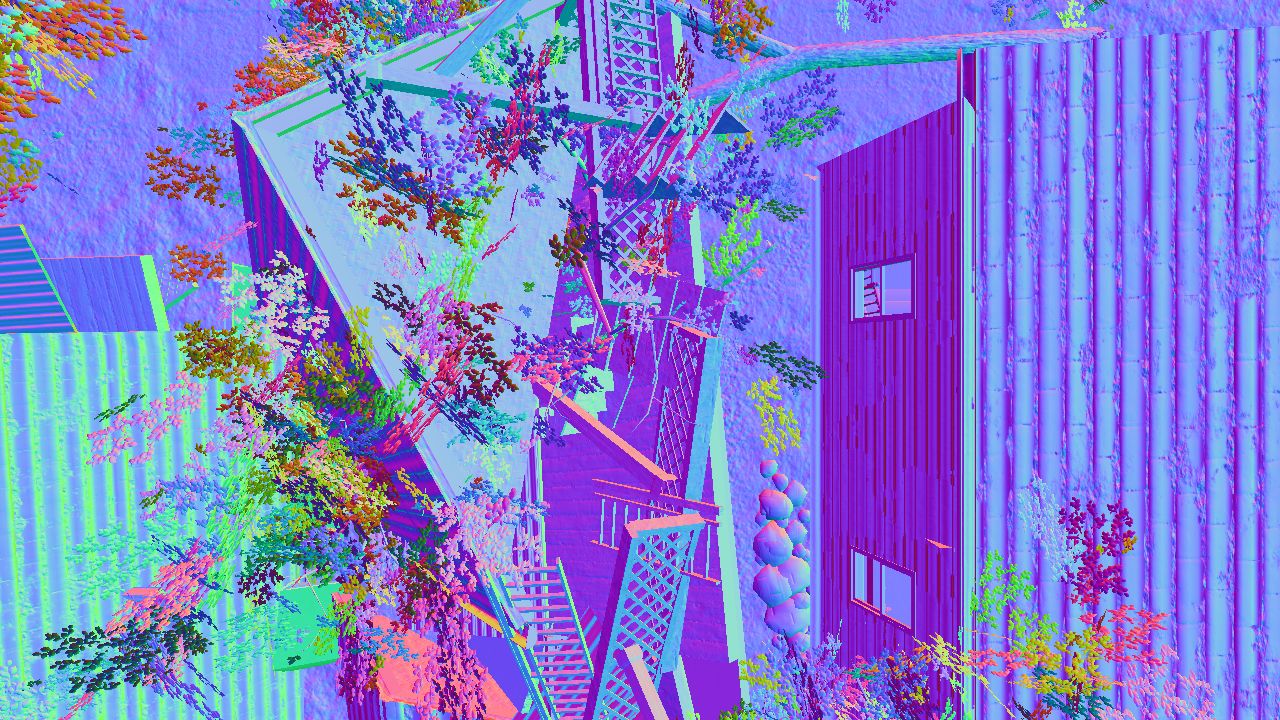} & 
    \includegraphics[width=\linewidth]{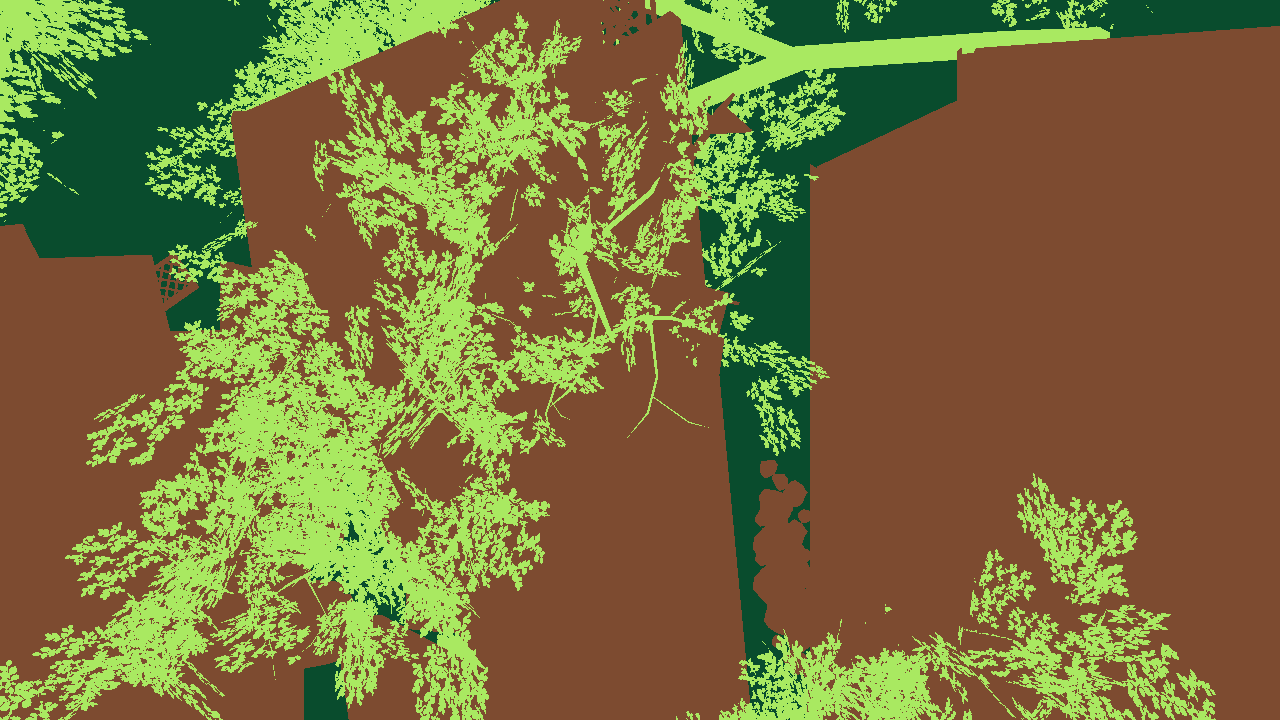} \\
\end{tabular}
\caption{Example images from our hyperrealistic synthetic dataset showing the RGB image with the corresponding groundtruth depth, surface normals and pixel-level semantic labels for nine object categories. Note that the depth image and the surface normals are colorized only for visualization.}
  \label{fig:airsim_sample}
  \vspace{-0.5cm}
\end{figure}

We use the open-source \textit{AirSim} plugin~\cite{airsim2017fsr} with our Unreal Engine environment and ROS as the middleware for our simulation experiments. The simulator considers forces such as drag, friction, and gravity in its physics engine, and provides data from various inertial sensors required for state estimation. We simulate a downward-facing stereo camera mounted on the UAV which provides RGB-D data at $20\hertz$ with a resolution of $640\times480$ pixels, similar to our real-world setup. As there are no publicly available datasets of collapsed buildings with groundtruth 6-DoF camera pose, depth and surface normals information, we collected a large-scale dataset from our simulation environment that we make publicly available\footnote{ Dataset publicly available at \url{http://autoland.cs.uni-freiburg.de}}. The dataset consists of 1,281,125 RGB images with the corresponding groundtruth for depth, surface normals, semantics and camera pose information. In order to have diverse viewing angles, we varied the tilt of the camera from $0\degree$ to $55\degree$ in steps of $5\degree$, the pan of the camera from $0\degree$ to $360\degree$ in steps of $45\degree$, and we also varied the height of the UAV during data collection from $10\meter$ to $30\meter$ in steps of $5\meter$. Additionally, we provide pixel-level groundtruth annotations for nine semantic object categories including Terrian, Trees, Flora, Sky, Roads, Rocks, Houses, Cars, and Other. Example images and the corresponding groundtruth from our dataset are shown in \figref{fig:airsim_sample}.

We use the same pipeline for state estimation, trajectory planning, and landing site detection as in our real-world UAV system. We build an Octomap at a resolution of $0.5\meter$ and the textured mesh at a resolution of $0.1\meter$. \figref{fig:results}(a) shows an example mesh visualization of a city block created using TSDFs while the UAV followed a lawn-mower surveillance path with a speed of $0.5\si[per-mode=symbol]{\meter\per\second}$. For the simulation experiments, we set the weights $c_1 = 0.05$, $c_2 = 0.4$, $c_3 = 0.4$ and $c_4 = 0.15$ for the depth accuracy, flatness, steepness and energy consumption costmaps respectively. We consider all the points with scores above $0.72$ in the final decision map as candidate landing sites. We set $c_1 = 0.05$ as the depth map is the groundtruth generated from the simulator. Since flatness and steepness play an equally important role in landing sites detection, we set their weights as equal. We set the euclidean distance and depth threshold for hierarchical clustering of landing sites to $0.50\meter$ and $0.01\meter$. 


\begin{figure}[t]
\centering
\footnotesize
\setlength{\tabcolsep}{0.1cm}
\begin{tabular}{P{3cm} P{4.1cm} P{4.1cm}}
\includegraphics[width=\linewidth]{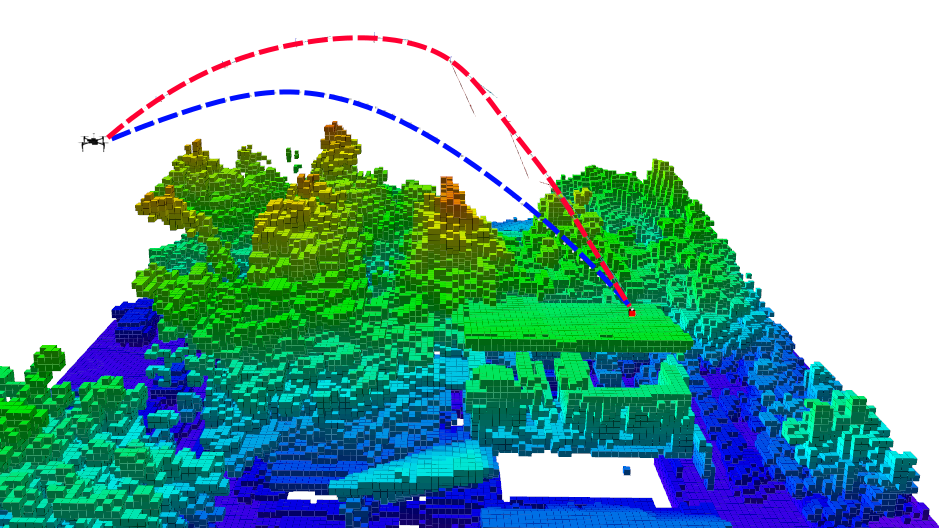} &
\includegraphics[width=\linewidth]{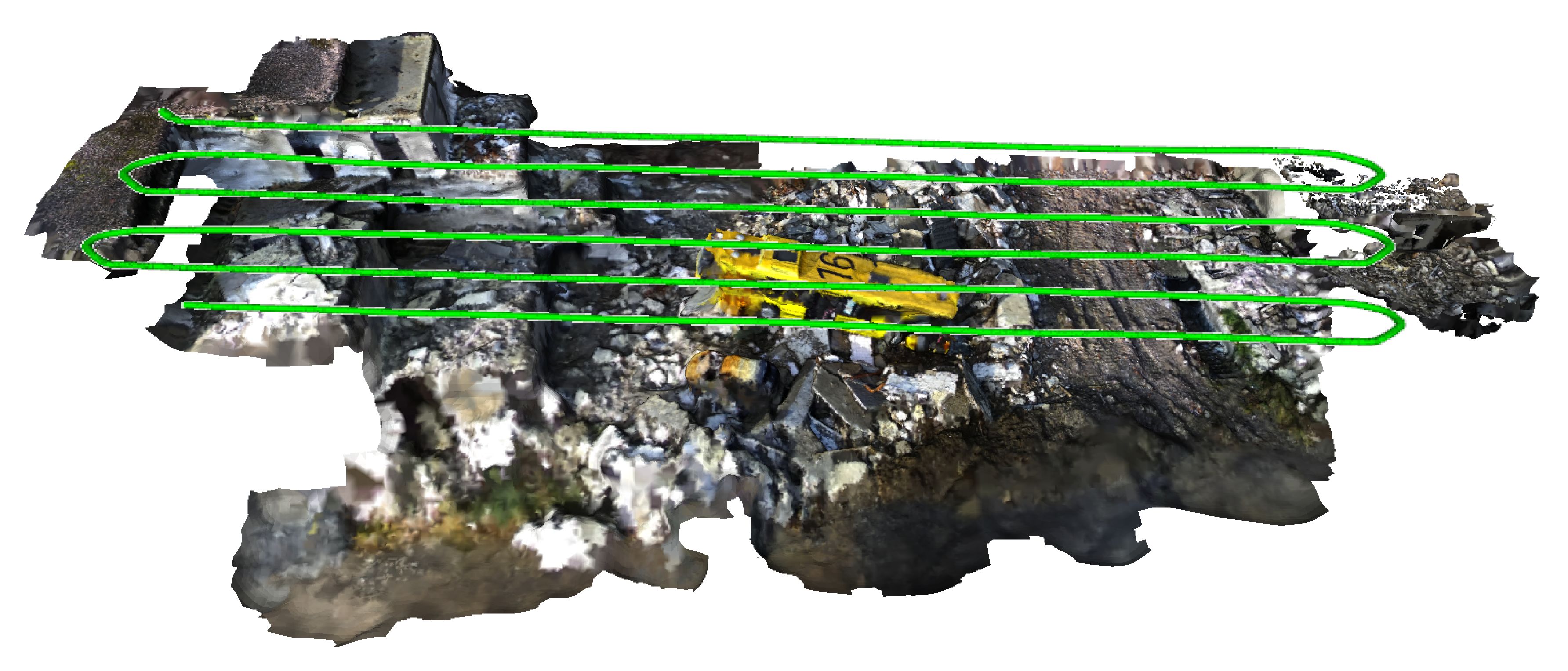} &
\includegraphics[width=\linewidth]{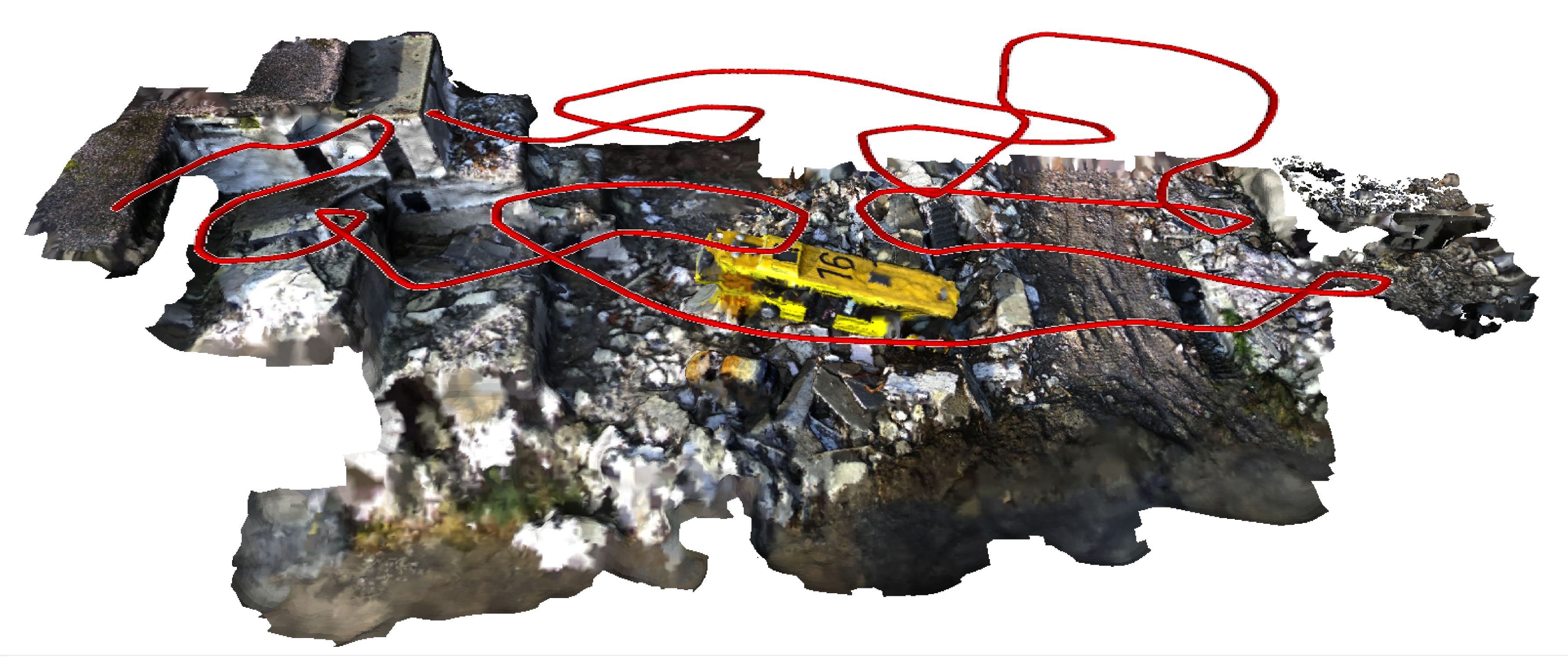} \\
(a) Landing trajectory & (b) Lawn-mower trajectory & (b) Next-best-view trajectory
\end{tabular}
\caption{\textit{(a)}: Comparison of the path generated using a strictly sampling-based RRT* approach with a polynomial steer function (blue line) and joint polynomial optimization method (red line). \textit{(b)}: Trajectory generated by the lawn-mower scan that was traversed by our UAV. \textit{(c)}: Trajectory generated by the next-best-view planner that traversed by our UAV.}
\label{fig:explorationTraj}
\end{figure}

\begin{figure}[t]
\centering
\footnotesize
\setlength{\tabcolsep}{0.35cm}
\begin{tabular}{P{5cm} P{5cm}}
\includegraphics[width=\linewidth]{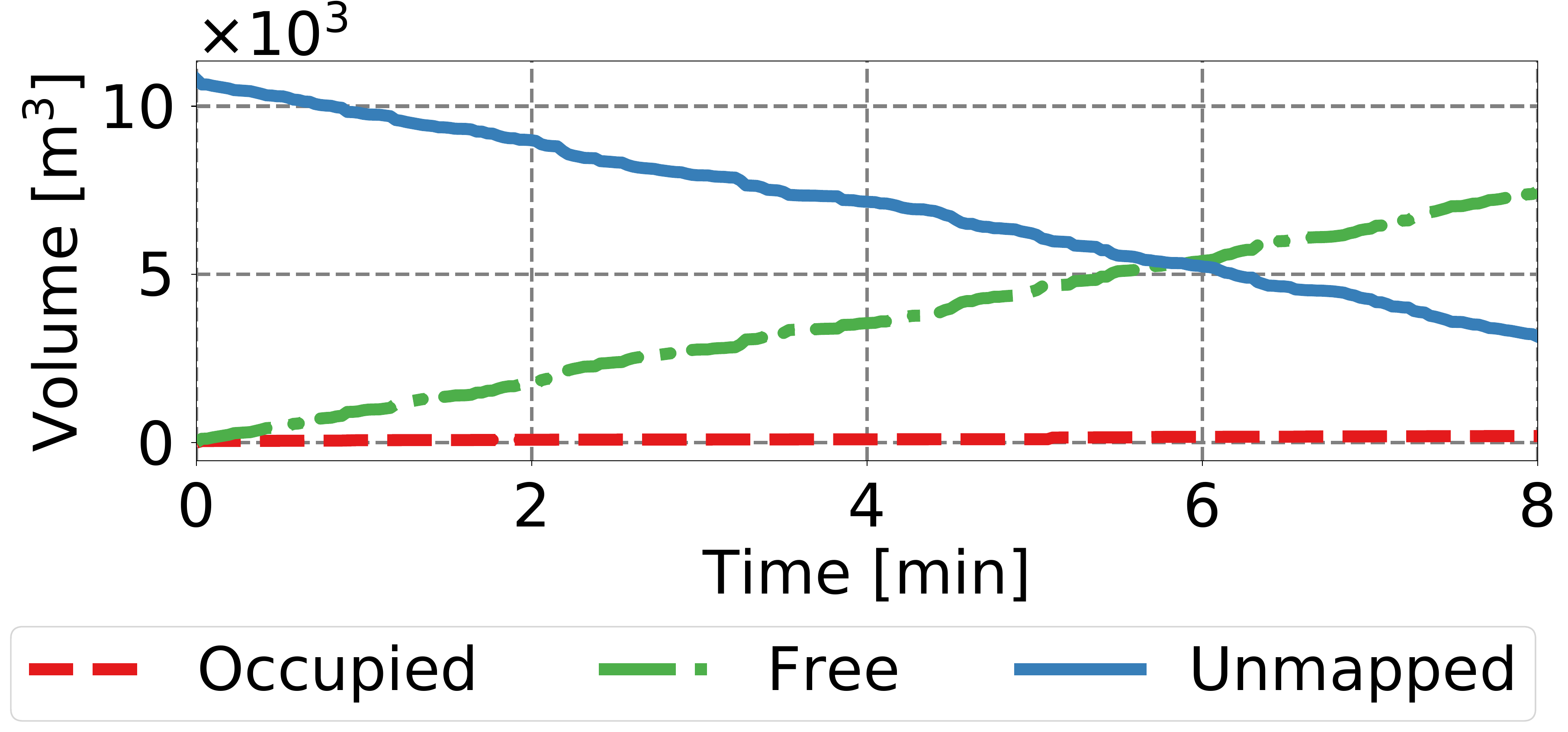} &
\includegraphics[width=\linewidth]{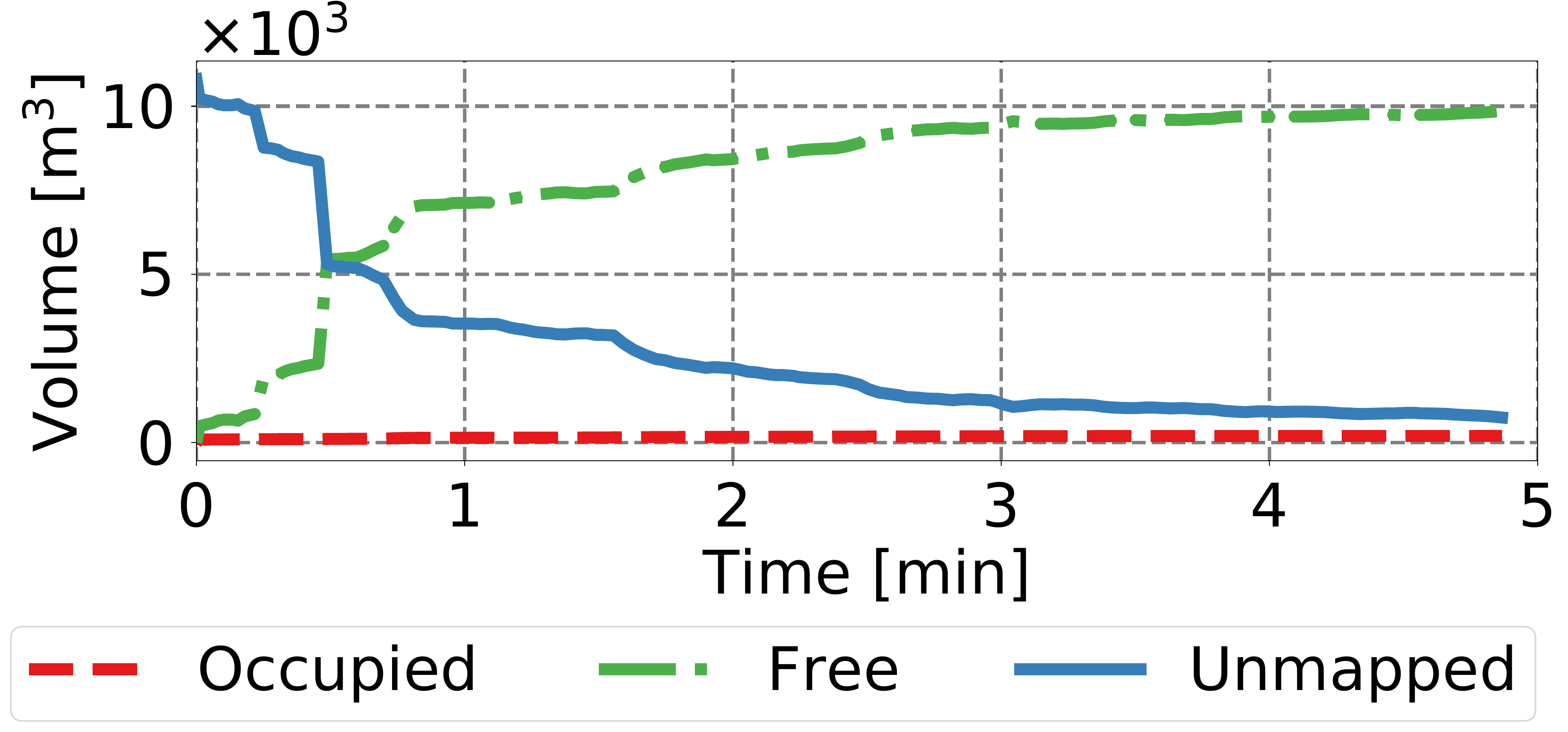} \\
(a) Lawn mower scan pattern & (b) Next-best-view planner
\end{tabular}
\caption{Comparison of exploration progress for the real-world scenario between the path generated from the next-best-view planner and the conventional lawn-mower scan pattern. The unmapped space decays at a faster rate using the next-best-view planner.}
\label{fig:exploration}
\vspace{-0.5cm}
\end{figure}

As it can be seen from scenarios $a.1$ and $a.3$ , the flatness costmap ensures that the detected landing sites are not close to the edges of the roof of broken buildings or other obstructions. On the other hand, in the scenarios $a.4$ and $a.5$, the steepness score ensures that steep flat surfaces such as sections of broken walls are avoided from being detected as landing sites. Moreover, in the scenarios $a.2$, $a.3$, and $a.5$, we observe that trees are automatically given a lower score in both the flatness and steepness costmaps. This discards them from possible landing site search area which is crucial in catastrophe struck environments as they often have uprooted trees. 

Fig.~\ref{fig:explorationTraj}(a) compares the trajectories generated using a strictly sampling-based RRT* approach and the joint polynomial optimization method used in our system. The landing site is selected from view $a.1$ in \figref{fig:results}. We observe that RRT* requires a much longer time to find a smooth path and cost of the path is higher.

\subsection{Experiments in Real-World Disaster Scenarios}

\begin{figure}[t]
\centering
\footnotesize
\begin{tabular}{P{11cm}}
\includegraphics[width=\linewidth]{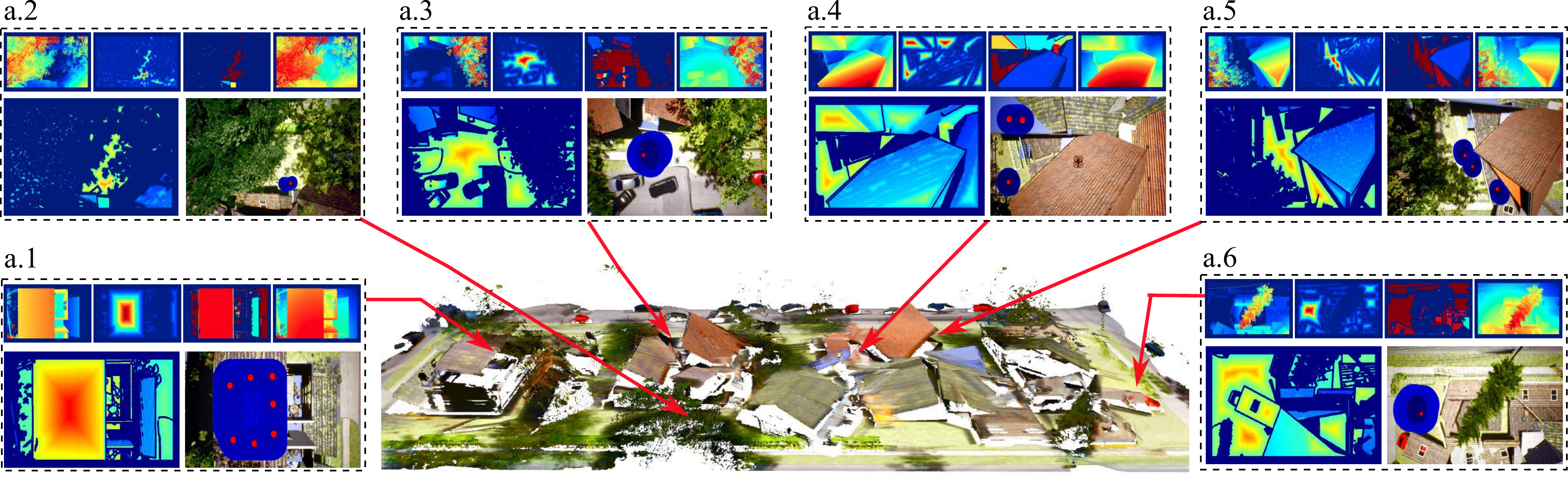} \\
(a) Hyperrealistic Simulation Environment  \\
\\
\includegraphics[width=\linewidth]{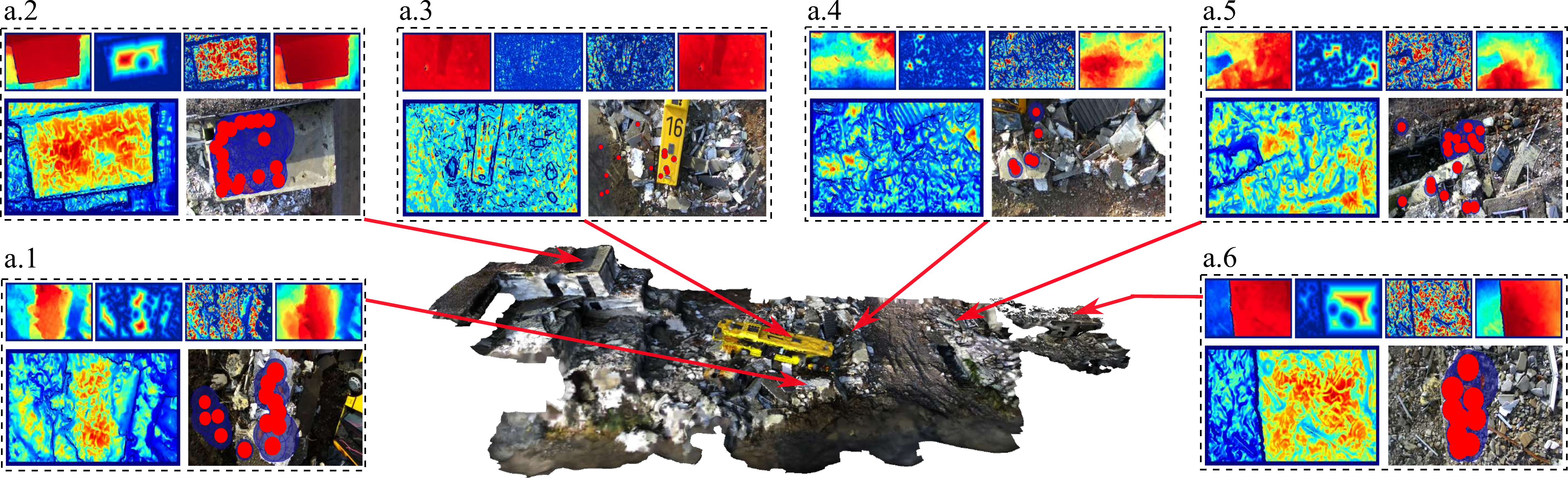} \\
(b) Real-World Outdoor Environment \\ 
\end{tabular}
\caption{Illustrations of costmap evaluation and dense landing site detection stages in both simulated and real-world scenarios. \textit{Inset:} The top row shows the depth accuracy, flatness, steepness and energy consumed costmaps respectively. The lower row shows the final decision map and the detected landing sites projected on to the camera image. The red circle denotes the projected size of the UAV at that location, while the blue circle is the inscribed circle generated during the evaluation of flatness costmap.}
\label{fig:results}
\vspace{-0.5cm}
\end{figure}

We use the DJI M100 quadrotor equipped with an NVIDIA TX2 embedded computer and a downward-facing ZED stereo camera for our real-world experiments. The stereo camera provides RGB-D images with a resolution of $640\times480$ pixels at $20\hertz$. Similar to the simulation experiments, we use ROS as the middleware on the TX2 which runs all the processes in real-time for state estimation, trajectory planning, landing site detection, and bio-radar processing. We use an Octomap resolution of $0.5\meter$ and a textured mesh resolution of $0.1\meter$. \figref{fig:results}(b) shows an example of mesh visualization created by the UAV while exploration. In \figref{fig:explorationTraj}(b) and (c) we show the paths followed by the UAV  while employing the next-best-view (NBV) scheme for exploration and while following a lawn-mower scan pattern respectively. The number of unmapped space decays faster when following the NBV scheme (\figref{fig:exploration}). This confirms that an NBV planner would ensure more efficient coverage of the catastrophe-struck environment which is important for such missions. 

Since in the real world, depthmaps generated from stereo images are often noisy, we set $c_1 = 0.15$ in order to account for this factor in the depth accuracy costmap. We set the weights for the flatness, steepness, and energy consumption costmaps to $c_2 = 0.35$, $c_3 = 0.4$ and $c_4 = 0.1$ respectively, while we set the overall thresholding value for the final decision map to $0.7$. For the hierarchical clustering parameters, we choose $0.5\meter$ for the euclidean distance (frame size of UAV being $0.26\meter$) and the depth threshold as $0.05\meter$. The effect of noisy depth data can also be seen in the other costmaps due to which the final decision map does not appear as clear as in the simulation environment. Nevertheless, our landing site search algorithm demonstrates detection of safe landing sites reliably even in situations where an expert safety operator is unable to make decisions.

In scenarios $b.1$, $b.4$ and $b.5$, landing sites are clearly detected on flat surfaces engulfed by debris from collapsed buildings. Similar to the results observed in the simulation environment, several landing sites are reliably detected on roofs of collapsed structures in the $b.1$ and $b.2$ scenarios. Moreover, landing sites are also detected on the roof of the bus in the $b.3$ scenario. While, in the $b.6$ scenario, landing sites are detected closer to the edge of the roof containing small rubbles which have tolerable roughness for landing. These sites are often safer to land on in comparison to other sites on the roof where larger rocks can be seen. 

\subsection{Comparitive Analysis of Landing Sites Detection}

\begin{figure}[t]
\centering
\footnotesize
\setlength{\tabcolsep}{0.15cm}
\begin{tabular}{P{2.8cm}P{2.3cm}P{2.8cm}P{2.4cm}}
Scenario 1 & Scenario 2 & Scenario 3 & Scenario 4 \\
\includegraphics[width=\linewidth]{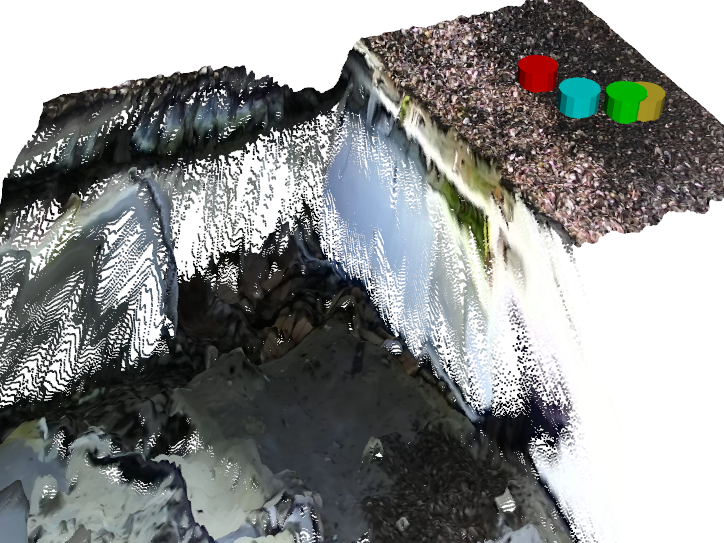} &
\includegraphics[width=\linewidth]{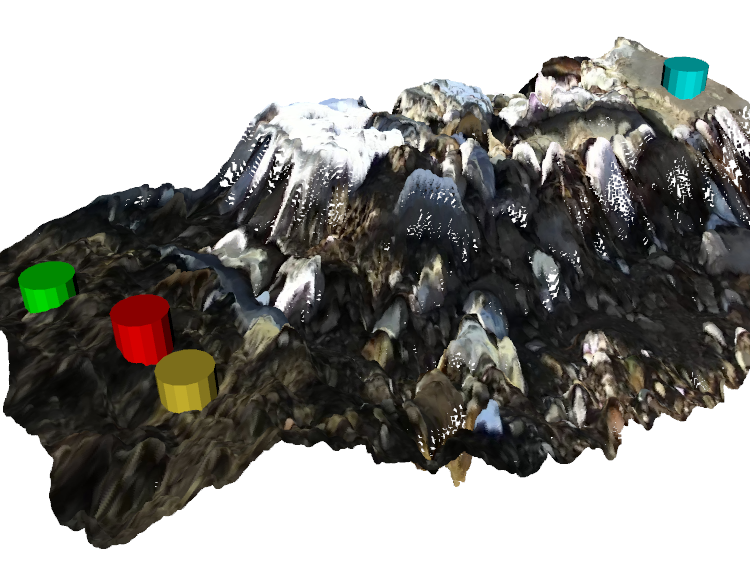} & \includegraphics[width=\linewidth]{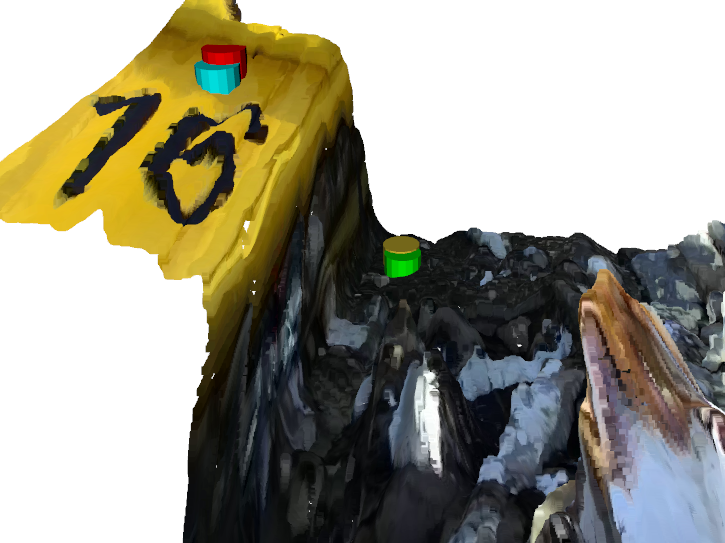} & \includegraphics[width=\linewidth]{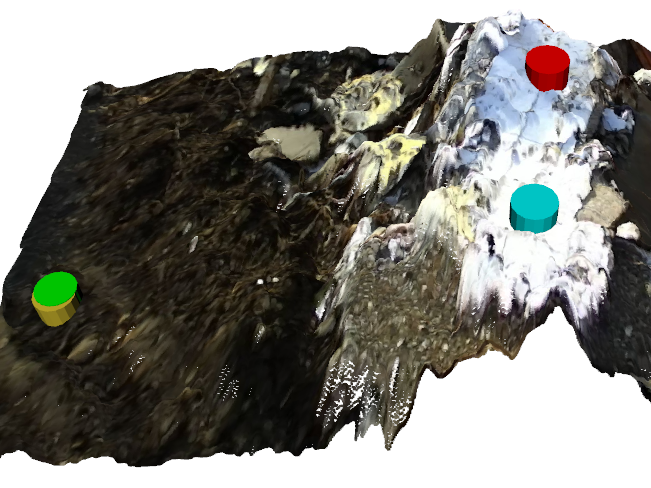}
\end{tabular}
\caption{Qualitative comparison of detected landing sites from existing algorithms and our proposed approach. For each algorithm, we only show the site with the highest score. \textit{Marker Legend:} Cyan - Ours, Red - Park \textit{et al.}~\cite{park2015landing}, Green - Forster \textit{et al.}~\cite{forster2015continuous}, Yellow: Hinzmann \textit{et al.}~\cite{hinzmann2018free}}
\label{fig:comparison_results}
\end{figure}

We compare our landing sites detection approach with respect to existing approaches proposed by Park \textit{et al.}~\cite{park2015landing}, Forster \textit{et al.}~\cite{forster2015continuous} and Hinzmann \textit{et al.}~\cite{hinzmann2018free}. We tune the hyperparameters for these algorithms on a subset of distinct regions in our dataset and use these values for comparisons on other scenes in the dataset. Results in~\figref{fig:comparison_results} shows a qualitative comparison of these three approaches and our proposed approach. We only show the landing site with the highest score that each of these approaches detects for a fair comparison. As it is seen from Scenario 1, all the algorithms perform well in regions which have a clear area like a rooftop to land on. This is expected since in these cases a simple metric such as terrain slopeness works well for landing site detection. However, in areas with a lot of debris (scenarios 2, 3, and 4), our algorithm picks sites on top of rubbles while the other algorithms tend to detect sites on homogenous regions like roads. While areas like roads are safe to land on, we want the UAV to land on collapsed buildings or debris so that the bioradars can detect any trapped victims under these structures. Our proposed algorithm performs better in such cases as the costmaps help to capture the fine-grained details of the terrain making our approach more robust and reliable for landing in catastrophe-struck environments. 

\begin{figure}[t]
\centering
\footnotesize
\setlength{\tabcolsep}{0.3cm}
\begin{tabular}{P{4cm} P{4cm} P{0.6cm}}
Case 1: Normal Operation & Case 2: Emergency Operation \\
\includegraphics[width=\linewidth]{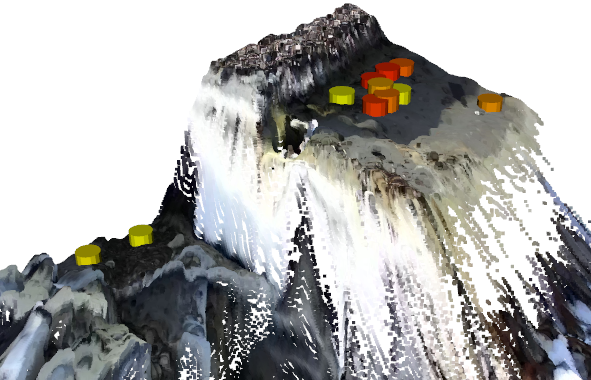} &
\includegraphics[width=\linewidth]{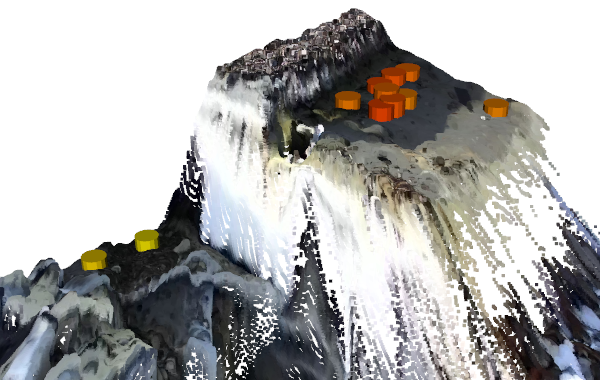} &
\includegraphics[height=2.5cm]{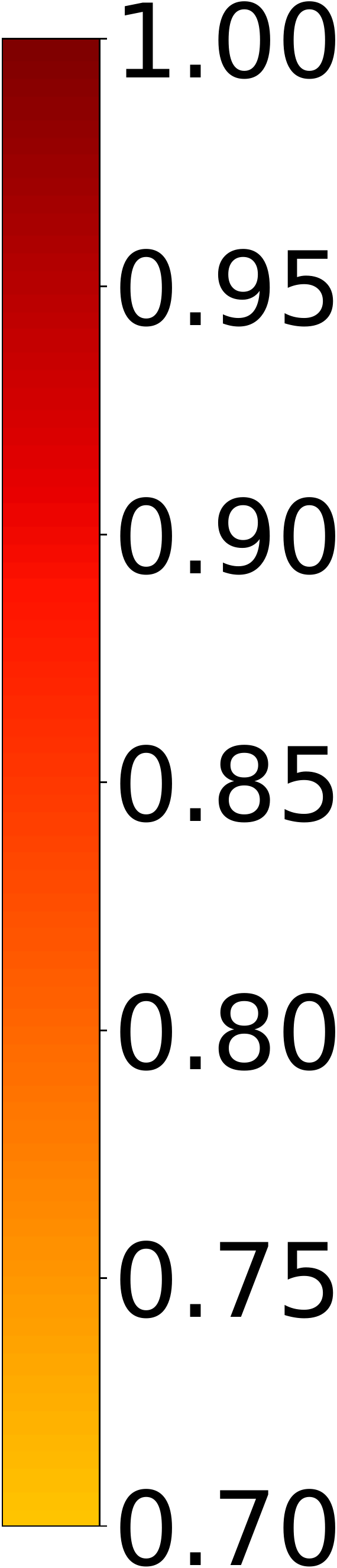}
\end{tabular}
\caption{Qualitative analysis of our proposed landing site detection algorithm for two different cases showing the dense sites. The color signifies the score from the final decision map $J$ assigned to each landing site. We set the threshold for a safe landing site as $0.7$. Case 1: During normal operation $(c_1, c_2, c_3, c_4) = (0.15, 0.35, 0.4, 0.1)$. Case 2: During immediate landing requirement (due to low battery) $(c_1, c_2, c_3, c_4) = (0.1, 0.1, 0.1, 0.7)$. }
\label{fig:ablative_results}
\vspace{-0.5cm}
\end{figure}

We further show through~\figref{fig:ablative_results} how changing the relative weighting between different costmaps can help during different requirements for the UAV. During normal operation, we keep the weights as $c_1 = 0.15$, $c_2 = 0.35$, $c_3 = 0.4$, and  $c_4=0.1$. This ensures that the UAV finds regions that are safe in terms of both the size and the terrain slope. However, when the UAV needs to land immediately, which may be triggered during situations such as a low battery, the weight to the energy costmap $c_4$ is increased since we want to retrieve landing sites that have a lower cost to go to. This flexibility provided by different weighting is incorporated into the high-level mission planner to handle the different needs of the UAV.

\subsection{Runtime Analysis}

Runtime efficiency is one of the critical requirements of our system as all the onboard processes for state estimation, planning, landing site detection, and bio-radar analysis have to run online on the embedded computer. The total computation time of the entire landing site detection algorithm, with the Octomap and textured mesh reconstruction along with the state estimation running in the backend, amounts to $173.4\milli\second$ on an Intel Core i7-8750H CPU @ 2.20GHz. A detailed breakdown of this computation time for each of the components of our system is shown in \tabref{tab:timing}.

\begin{table}[t]
\centering
\footnotesize
\setlength\tabcolsep{0.5cm}
\caption{Computational runtime analysis of our landing sites detection system. Plot on the right top shows the time consumed to evaluate individual costmaps for each frame and the plot on the right bottom shows the time consumed by each of the stages of the landing site detection system.}
\vspace{5pt}
\label{tab:timing}
\newcolumntype{d}[1]{D{?}{\pm}{#1}}
\begin{tabular}{@{}p{2.5cm} d{1.1}}
\toprule
\textbf{Algorithm} &  \multicolumn{1}{@{}p{2.1cm}}{\textbf{Time ($\mu \pm \sigma$) $\milli\second$}} \\
\noalign{\smallskip}\hline\hline\noalign{\smallskip}
   Costmap Evaluation &  \\
   \tab Depth Accuracy & 5.1 ? 0.2 \\
   \tab Flatness &  91.7 ? 38.8 \\
   \tab Steepness & 25.1 ? 4.2 \\
   \tab Energy & 5.1 ? 0.6 \\
   \tab Final & 5.1 ? 0.3 \\
   Dense Detection & 11.0 ?  14.9 \\
   Clustering & 125.5 ? 100.3 \\
\midrule
   \textbf{Total Time} & \textbf{173.4} ? \textbf{59.2} \\
\bottomrule
\end{tabular}
\hspace{0.15cm}
\begin{tabular}{P{4.8cm}}
\includegraphics[width=\linewidth]{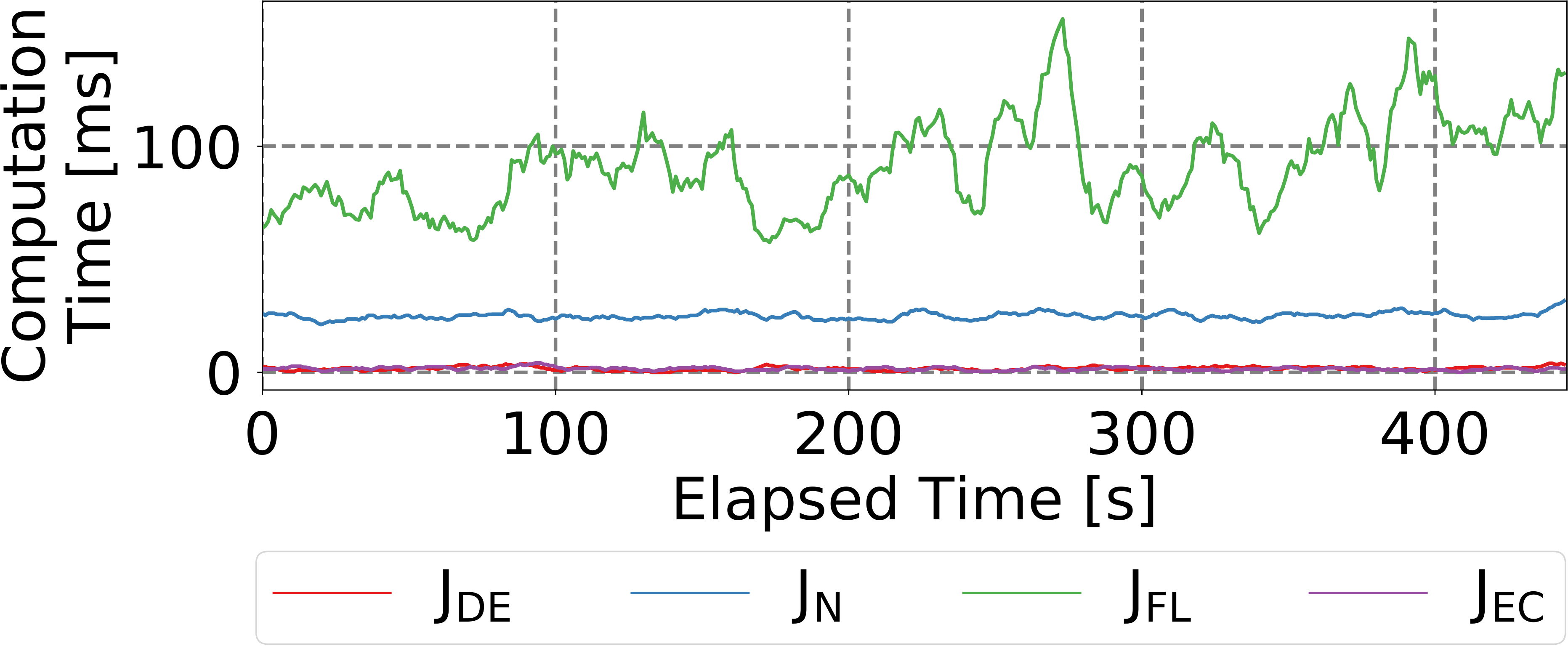} \\
\includegraphics[width=\linewidth]{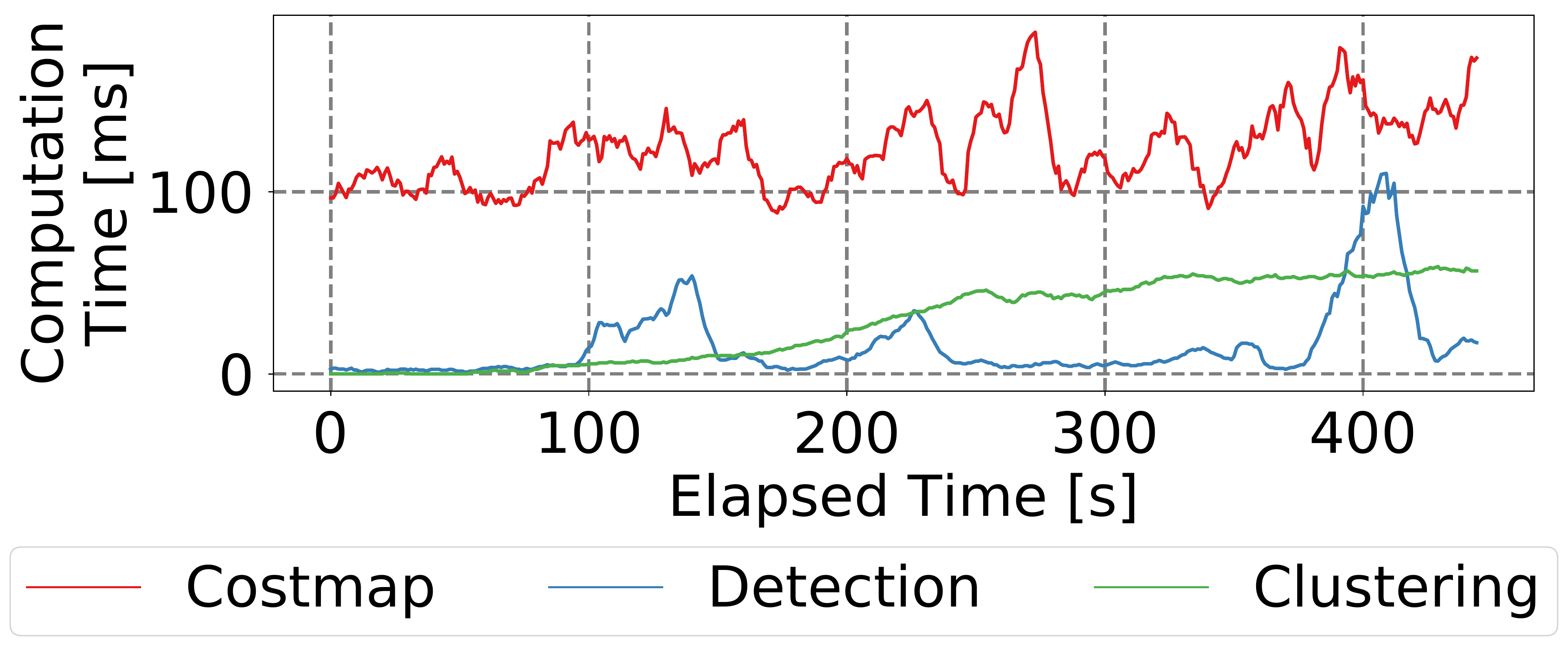} \\
\end{tabular}
\vspace{-0.5cm}
\end{table}

The computation time for costmaps was evaluated for each input depth map. The cost of computing the depth accuracy $J_{DE}$, steepness $J_N$ and energy consumption $J_{EC}$ scores linearly depends on the size of the input image. During operation, the resolution of sensor images is unchanged leading to a constant computation time to evaluate these costmaps. On the other hand, the computation time consumed for evaluating the flatness score $J_{FL}$ depends on the distance transformation operation which varies with the image content of the binary map obtained from canny edge operation. Due to this factor, a large variation in the computation time is observed while evaluating the flatness costmap. The dense landing sites detection step involves identifying candidate landing sites on the basis of their scores in the final decision map and aggregating these sites into a global list using k-d trees while removing duplicates. Since the number of landing sites detected in each frame varies according to the scene, we observe a large variance in the computation time for this step. Similarly, as the number of total landing sites detected increases, the time taken by the clustering step increases to a certain extent but then eventually remains constant, therefore, it does not affect the real-world operation.

\section{Conclusion}
\label{sec:conclusion}

In this paper, we presented a vision-based autonomy system for UAVs equipped with bioradars tasked to assist in USAR operations in post-disaster struck environments. We demonstrated how recent developments in the field of robotics including state estimation, mapping, and high-level trajectory planning can be integrated to achieve an efficient and reliable system for time-critical tasks during a disaster management cycle. Using a next-best-view planner and a TSDF-based reconstruction algorithm, our UAV swiftly builds a three-dimensional textured mesh-representation of the environment to assist the first response team in structural analysis and rescue planning.

We presented a novel landing site detection algorithm that enables UAVs to safely land on top of rubble piles and debris from collapsed buildings to locate trapped victims using the bioradar. Our algorithm weighs several hazard factors such as the flatness and steepness of the terrain, accuracy of the depth measurements, and the energy required for landing, to detect dense candidate landing sites. It employs nearest neighbor filtering and clustering to group these dense sites into safe landing regions. Our system also computes a minimum-jerk trajectory to the landing site considering nearby obstacles as well as the UAV dynamics to safely execute the landing maneuver. The proposed system is computationally efficient as it runs online on an onboard embedded computer. We also introduced a first-of-a-kind synthetic dataset of an environment affected by a natural disaster, consisting of over 1.2 Million RGB images with corresponding groundtruth depth, surface normals, pixel-level semantic segmentation labels and camera pose information. We demonstrated the utility of our system using extensive experiments on our synthetic dataset and in real-world environments representing catastrophe struck scenarios caused by earthquakes or gas explosions.

\begin{acknowledgement}
This work has partly been supported by the Federal Ministry of Education and Research
of Germany through the project FOUNT2.
\end{acknowledgement}

\bibliographystyle{spmpsci}
\bibliography{sections/references}
\end{document}